\documentclass{article}

\usepackage{microtype}
\usepackage{graphicx}
\usepackage{booktabs,colortbl} 

\usepackage{hyperref}

\usepackage[accepted]{icml2021}

\usepackage{todonotes}

\usepackage{url}
\usepackage{amsfonts}
\usepackage{amssymb}
\usepackage{amsmath}
\usepackage{pifont}
\usepackage{bm}

\usepackage{caption}
\usepackage{subcaption}
\usepackage{multirow}
\usepackage{diagbox}

\usepackage{enumerate}
\usepackage{arydshln}
\usepackage{tablefootnote}
\usepackage{footmisc}
\usepackage{placeins}
\usepackage{tikz}
\usetikzlibrary{positioning, arrows.meta, shapes.geometric}
\usepackage[noabbrev, capitalise]{cleveref}

\usepackage{siunitx}
\def\eqref#1{equation~\ref{#1}}

\def\1{\bm{1}}

\def\vh{{\bm{h}}}

\def\vr{{\bm{r}}}

\def\vt{{\bm{t}}}

\DeclareMathAlphabet{\mathsfit}{\encodingdefault}{\sfdefault}{m}{sl}
\SetMathAlphabet{\mathsfit}{bold}{\encodingdefault}{\sfdefault}{bx}{n}

\usepackage{pifont}

\newcommand{\R}{\mathbb{R}}
\newcommand{\C}{\mathbb{C}}

\newcommand{\norm}[1]{\left\lVert#1\right\rVert}

\newcommand{\cmark}{\ding{51}}
\newcommand{\xmark}{\ding{55}}

\icmltitlerunning{The Role of Graph Topology in the Performance of Biomedical Knowledge Graph Completion Models}

\begin{document}

\twocolumn[
    \icmltitle{The Role of Graph Topology in the Performance of Biomedical Knowledge Graph Completion Models}




    \begin{icmlauthorlist}
        \icmlauthor{Alberto Cattaneo}{gc}
        \icmlauthor{Stephen Bonner}{az}
        \icmlauthor{Thomas Martynec}{az}
        \icmlauthor{Edward Morrissey}{az}
        \icmlauthor{Carlo Luschi}{gc}
        \icmlauthor{Ian P.\ Barrett}{az}
        \icmlauthor{Daniel Justus}{gc}
    \end{icmlauthorlist}

    \icmlaffiliation{gc}{Graphcore Research, Bristol, UK}
    \icmlaffiliation{az}{Data Sciences and Quantitative Biology, Discovery Sciences, R\&D, AstraZeneca, Cambridge, UK}

    \icmlcorrespondingauthor{Alberto Cattaneo}{albertoc@graphcore.ai}
    \icmlcorrespondingauthor{Daniel Justus}{danielj@graphcore.ai}

    \icmlkeywords{Machine Learning, Knowledge Graphs, Knowledge Graph Completion, Graph Topology, Drug Discovery, Biomedical Research}

    \vskip 0.3in
]



\printAffiliationsAndNotice{}  

\begin{abstract}
    Knowledge Graph Completion has been increasingly adopted as a useful method for helping address several tasks in biomedical research, such as drug repurposing or drug-target identification.
    To that end, a variety of datasets and Knowledge Graph Embedding models have been proposed over the years.
    However, little is known about the properties that render a dataset, and associated modelling choices, useful for a given task. Moreover, even though theoretical properties of Knowledge Graph Embedding models are well understood, their practical utility in this field remains controversial.
    In this work, we conduct a comprehensive investigation into the topological properties of publicly available biomedical Knowledge Graphs and establish links to the accuracy observed in real-world tasks.
    By releasing all model predictions and a new suite of analysis tools we invite the community to build upon our work and continue improving the understanding of these crucial applications.
\end{abstract}

\section{Introduction}\label{sec:introduction}

Knowledge Graphs (KGs) have become a powerful way of integrating information about a domain into a single structured representation.
KGs represent knowledge as a collection of facts in the form of triples \((h,r,t)\), which make a statement about the relationship \(r\) between two entities \(h\) and \(t\).
In KGs from the biomedical domain, entities can be genes, diseases, drugs, or pathways (among others), whilst relations represent biological interactions or associations between them~\citep{maclean2021knowledge, bonner2022_review, zeng2022toward}.
Biomedical KGs have been used to support a variety of tasks within the complex and costly drug discovery process, such as drug repurposing~\citep{bang2023biomedical}, identifying synergistic drug combinations~\citep{zitnik2018_modeling} and target discovery~\citep{paliwal2020_preclinical}.
Unlike in other domains, biomedical KGs frequently comprise information from a wide range of conceptual abstraction levels, for both entities and edges.
It is common to find edges representing information with a low abstraction level, such as experimentally validated protein-protein interactions (\emph{Protein A, interacts, Protein B}), in the same KG as mid abstraction level edges, such as associations between genes and bioprocesses (\emph{Gene A, associates, BioProcess X}), and high abstraction level edges, such as disease ontology hierarchies (\emph{Disease X, subclass of, Disease Y}).
Entities can also have different abstraction levels, with some representing physical entities (e.g., genes) and others representing conceptual entities (e.g., diseases or biological processes).
This makes biomedical KGs uniquely interesting to consider when they are used in combination with Machine Learning algorithms, as models must learn to make predictions across these different abstraction levels simultaneously.

Since KGs, including biomedical ones, are typically incomplete, inferring missing triples is a key Machine Learning application -- a task known as Knowledge Graph Completion in the literature.
Perhaps the most common method of making predictions on KGs involves using Knowledge Graph Embedding (KGE) models, which learn low dimensional representations for all entities and relation types in a KG and combine them through a scoring function to compute the likelihood of a missing edge.
In the biomedical domain, Knowledge Graph Completion has been used for a diverse set of tasks including predicting new links between genes and diseases~\cite{paliwal2020_preclinical} and possible drug interactions, among others~\cite{gema2024knowledge}.

Biomedical KGs are often constructed by practitioners to address a specific purpose (e.g., investigate the pathways of a given disease), requiring decisions about the key entities and relation types to include, the optimal data sources from which to extract information and how best to model the desired knowledge in a KG schema.
This manual graph composition process is costly and time-consuming, an issue that can be compounded when KGs are designed as input for ML models, often requiring several iterations to best tune the graph to the predictive task at hand~\citep{bonner2022_review, ratajczak2022task}.
A solid understanding of desirable topological properties of KGs that can be leveraged by ML models to make better predictions would allow for a more principled approach to this graph construction process, making it cheaper and faster to generate KGs with higher utility to the practitioner.

Thus, there is a growing interest in understanding how any decisions taken which affect topological properties of a KG can ultimately impact predictions made from them.
Amongst other topological properties, patterns such as \emph{compositions} (\cref{fig:relational_patterns} (d)), where a relation between two entities is inferred by the presence of a connecting third entity, can be used to link graph structure to model prediction quality.
In the biomedical domain, this could be the case when a drug is inferred to treat a disease because they share a common gene connection. Ideally, we would like to discover new drug repurposing opportunities by learning from such patterns in the data.
Similarly, the \emph{inference} pattern (\cref{fig:relational_patterns} (b)) uses the presence of a certain relation type between an entity pair to infer the presence of a different relation between the same pair, which could, for example, be used to discover novel therapeutic gene-disease connections by aggregating multiple evidence types.
Another example is in Protein-Protein Interactions (PPI), where an edge could have a clear directionality if representing information flow in a pathway, or be inherently \emph{symmetrical} (\cref{fig:relational_patterns} (a)) if representing simple binary physical interaction.

While the theoretical implications of these topological patterns for KGE models are well understood -- in fact, new KGE models are often designed to capture key topological patterns -- their impact on the predictive performance in real-world Knowledge Graph Completion tasks remains an open question.
Recent work has started to investigate this~\citep{ali2021bringing, jin2023comprehensive, teneva2023_knowledge} but has primarily focussed on the properties of relation types rather than individual triples.
This approach can help to understand which topological patterns can be captured by the relation embedding when comparing or proposing new KGE models. However, when investigating the impact of the graph structure on KGE model performance, this relation-level approach might neglect critical local properties of the graph.
Moreover, existing work has focussed on a set of general-domain KGs, which may contain distinct topological patterns compared to biomedical KGs.

Thus, despite the clear need for understanding the relationship between the topology of KGs and model performance specifically in the biomedical domain, there is a gap in the existing literature, which we aim to address in this work by exploring such relationship at an unprecedented level of detail.

Our main contributions are as follows:
\begin{itemize}
    \item We provide an in-depth analysis of the topological properties of six public KGs, focussing on the biomedical domain, and compare the corresponding predictive performance of five well-established KGE models. While previous work has limited the investigation to the macro-level of relation types, we are able to detect stronger patterns linking topological properties to predictive accuracy by zooming in at the level of individual triples.
    \item We look in detail at highly relevant relations that biomedical practitioners are most interested in inferring. We give evidence of the topological differences in how they are represented in different KGs and how this reflects on predictive performance.
    \item To aid future work in this area, we propose a standardized framework to describe KG topology and release a dedicated Python package \texttt{kg-topology-toolbox}\footnote{\url{https://github.com/graphcore-research/kg-topology-toolbox}} offering a wide range of tools to compute topological statistics of any KG. To help bridge the gap between industry and academia, we also make the predictions of the trained models available to the community~\citep{zenodo_data} to conduct further analysis.
\end{itemize}
\section{Related Work}\label{sec:related_work}

Since the adoption of representation learning for Knowledge Graph Completion, there has been interest in linking predictive performance of models to topological structures in the underlying Knowledge Graphs~\citep{bordes2013_transe}. These structures, taking into account the heterogeneous relation types in KGs, have been split into two categories: edge cardinalities and topological patterns. For the definitions used in this study see section~\nameref{sec:topology}. Indeed, it has become common for new models to be designed with the capacity of capturing these patterns~\citep{cao2021_dual, chao2020_pairre, sun2019_rotate}, working under the assumption that better modelling them will result in improved predictive performance~\citep{jin2023comprehensive}.

Edge cardinality patterns group relations based on the cardinalities of their head and tail entity sets. Four cardinality patterns have been proposed: one-to-one, one-to-many, many-to-one, and many-to-many~\citep{bordes2013_transe, wang2014_knowledge, yang2015_embedding}. On the other hand, topological patterns categorise relationships based on underlying logical principles that define how entities interact. Several such patterns have been proposed, with common examples being symmetric/anti-symmetric, inverse and composition~\citep{sun2019_rotate, ali2021bringing}. In the literature, it is common to consider these patterns at the level of whole relation types, meaning that all triples with a given relation type must conform to a certain pattern for it to be said to be present.

Whilst both pattern types are frequently discussed when comparing architectures, quantifying how model predictive performance correlates with them is less common. Existing work has typically been limited in scale, often only performed on a new approach when it is proposed. For example, the original TransE paper investigated the impact of cardinality patterns on model performance, finding that models could more easily predict heads in one-to-many relations and tails in many-to-one relations~\citep{bordes2013_transe}. The DistMult model observed similar results, whilst also measuring the ability of the model to capture composition relations~\citep{yang2015_embedding}. TransH was found to be better at capturing symmetric relations~\citep{wang2014_knowledge}. RotatE demonstrated better predictive performance than existing models on composition relation types, whilst the authors also found implicit evidence of other topological patterns in the embeddings~\citep{sun2019_rotate}.

There has been limited work providing a comprehensive study on the impact of these patterns across different models and datasets, although some studies have started to appear recently. For example, an investigation across 3 common benchmark KGs (FB15k-237, WN18RR and Yago310) and 21 KGE models found that symmetric relations were easier to predict that anti-symmetric and compositions ones~\cite{ali2021bringing}. Another study compared the performance of 7 KGE models on FB15k-237 and WN18RR, focussing on 4 different topological patterns~\citep{jin2023comprehensive}, finding that a model's theoretical capacity to capture a certain pattern does not guarantee good predictive performance on it. Taking a different approach, a study showed how adding noise to a graph impacted the ability of TransE to learn the composition pattern~\citep{douglas2022learned}. However, none of these studies explores the impact of edge patterns on biomedical KGs specifically.

Biomedical KGs can suffer from degree imbalance, which can arise naturally from the biology modelled by the KG (e.g., key hub-genes having a high degree) or artificially (e.g., due to inspection bias where more attention is paid to high degree entities). This degree imbalance can impact downstream models: through perturbation of graph topology, degree alone has been shown to be a strong prior predictor for the existence of edges~\cite{zietz2024probability}. KGE models are known to be strongly impacted by entity degree, with models often being heavily biased towards highly connected entities~\cite{bonner2022_implications, rossi2021knowledge, rossi2021knowledgestudy, mohamed2020popularity}. However, disentangling the effect of degree from other topological properties when considering model performance is often overlooked by prior studies. One paper showed that as entity degree increases, performance on symmetric relations decreases~\citep{jin2023comprehensive}. However, it is still an open research question as to how degree obscures other correlations in the performance of models on different topological patterns.

The prevalence of different topological patterns in real-world datasets is also an active area of research. The frequency of topological and cardinality patterns in 29 real-world KGs from a range of domains, including biomedical was investigated in \citep{teneva2023_knowledge}. The authors observed that, under their chosen pattern definitions, many relations in biomedical KGs are predominantly many-to-many and that there was a lack of composition relations. Other work investigated the prevalence of symmetric and composition relations in 3 common KG benchmark datasets, finding that the majority of relations were anti-symmetric and that composition relations were rarer~\citep{ali2021bringing}.

It is also crucial to understand how the presence of topological patterns could impact validation metrics if not considered when creating train/test splits. The benchmark dataset FB15k was found to contain inverse triple pairs where one part was in the train set and its counterpart in the test set, meaning that models could simply memorise pairs~\citep{toutanova2015_fb15k-237}. A new dataset, FB15k-237, was created to remove these train/test set leakages. Another common dataset, WN18, was found to have a similar problem, with the WN18RR dataset created to address this~\citep{toutanova2015_fb15k-237}.
\section{Knowledge Graph Topological Properties}\label{sec:topology}
\subsection{Topological Patterns}

In terms of describing KG topology, we focus on both the cardinality of the head and tail of triples, as well as their edge topological patterns. Specifically, for a triple \((h, r, t)\) in a KG \(\mathcal{G}\) we define the \textit{head out-degree} \( \textrm{deg}(h) \) as the cardinality of the set \( \{\hat{t} \mid \exists \hat{r}: (h, \hat{r}, \hat{t}) \in \mathcal{G}\} \) and, analogously, the \textit{tail in-degree} \( \textrm{deg}(t) \) as the cardinality of \( \{\hat{h} \mid \exists \hat{r}: (\hat{h}, \hat{r}, t) \in \mathcal{G}\} \).
We further define the \textit{head out-degree of same relation type} \( \textrm{deg}_r(h) \) and the \textit{tail in-degree of same relation type} \( \textrm{deg}_r(t) \) as the cardinalities of \( \{\hat{t} \mid (h, r, \hat{t}) \in \mathcal{G}\} \) and \( \{\hat{h} \mid (\hat{h}, r, t) \in \mathcal{G}\} \), respectively.

\begin{figure}[h]
    \centering
    \includegraphics[width=.55\linewidth]{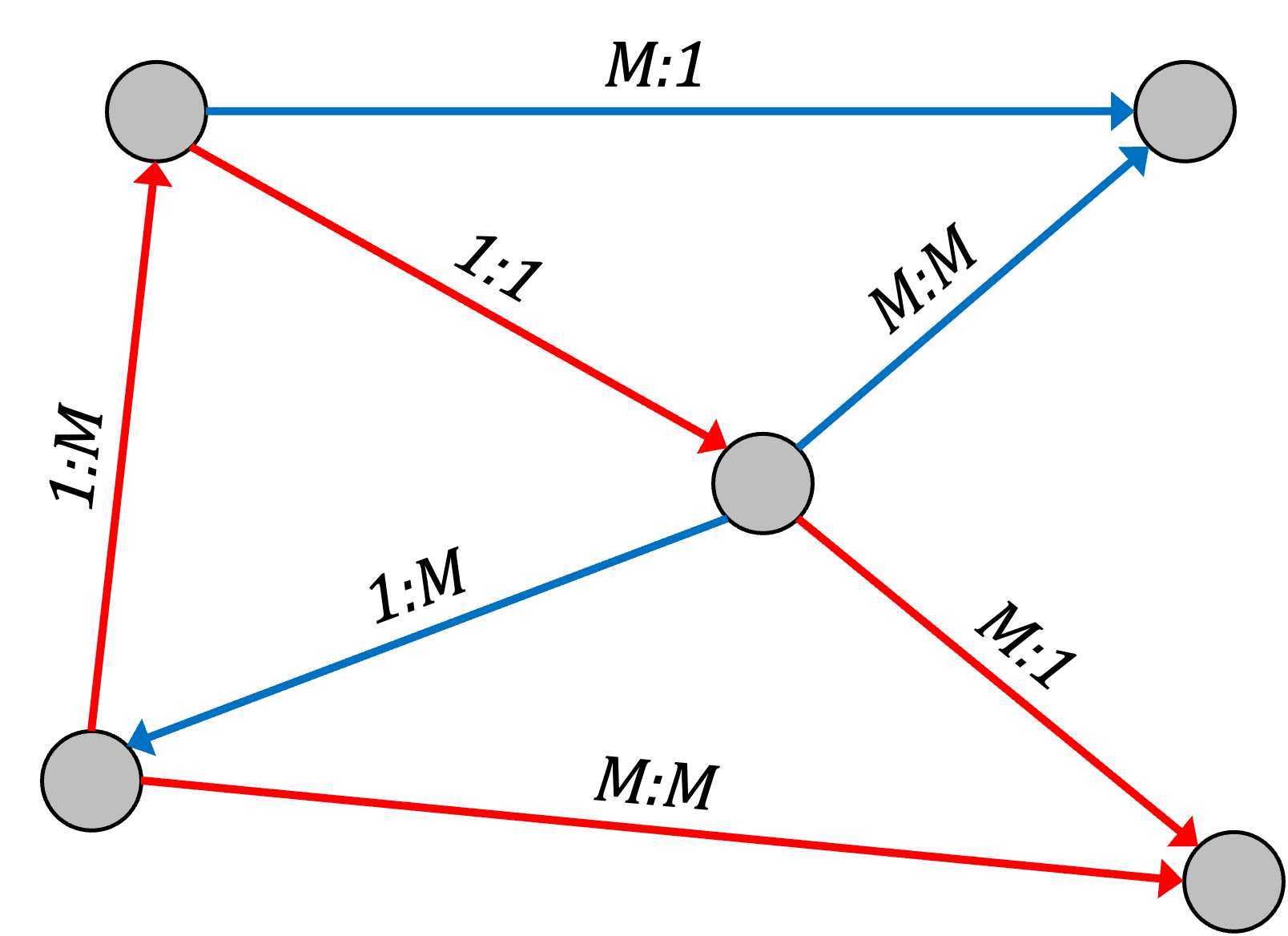}
    \caption{Example of edge cardinalities in a KG with two relation types (blue, red). We use the notation \( \textrm{deg}_r(t) \):\( \textrm{deg}_r(h) \). For example, for the red edge with cardinality 1:M, the head has more than one outgoing edges of the same relation type (red) and the tail has a single red incoming edge.}\label{fig:example_cardinalities}
\end{figure}

The \textit{edge cardinality} of a triple \( (h, r, t) \) is then defined as:

\begin{center}
    \resizebox{\columnwidth}{!}{%
        \begin{tabular}{|l|*{2}{c|}}\hline
            \backslashbox{\( \textrm{deg}_r(h) \)}{\( \textrm{deg}_r(t) \)}
                      & \( = 1 \)                  & \( > 1 \)                   \\\hline
            \( = 1 \) & \textit{one-to-one (1:1)}  & \textit{many-to-one (M:1)}  \\\hline
            \( > 1 \) & \textit{one-to-many (1:M)} & \textit{many-to-many (M:M)} \\\hline
        \end{tabular}}
\end{center}
Examples of the four possible edge cardinalities are provided using a small sample graph in \cref{fig:example_cardinalities}.

\begin{figure}[htb]
    \centering
    \subfloat[][Symmetry\label{fig:symmetric}]{
        \begin{tikzpicture}[node distance=1.1cm, auto,
                main node/.style={circle, draw, minimum size=0.65cm, fill=blue!15},
                label node/.style={fill=white},
                line/.style={-Latex, line width=0.4mm}]
            \node[main node] (h) {h};
            \node[main node, right=of h] (t) {t};
            \draw[line] (h) to[bend right=20]  node[midway, below] {$r$} (t);
            \draw[line, dashed] (t) to[bend right=20] node[midway, above] {$r$} (h);
        \end{tikzpicture}
    }
    \hfill
    \subfloat[][Inference\label{fig:inference}]{
        \begin{tikzpicture}[node distance=1.1cm, auto,
                main node/.style={circle, draw, minimum size=0.65cm, fill=blue!15},
                label node/.style={fill=white},
                line/.style={-Latex, line width=0.4mm}]

            \node[main node] (h) {h};
            \node[main node, right=of h] (t) {t};
            \draw[line, dashed] (h) to[bend left=20]  node[midway, above] {$r^\prime$} (t);
            \draw[line] (h) to[bend right=20] node[midway, below] {$r$} (t);
        \end{tikzpicture}
    }
    \hfill
    \subfloat[][Inverse\label{fig:inverse}]{
        \begin{tikzpicture}[node distance=1.1cm, auto,
                main node/.style={circle, draw, minimum size=0.65cm, fill=blue!15},
                label node/.style={fill=white},
                line/.style={-Latex, line width=0.4mm}]
            \node[main node] (h) {h};
            \node[main node, right=of h] (t) {t};

            \draw[line] (h) to[bend right=20]  node[midway, below] {$r$} (t);
            \draw[line, dashed] (t) to[bend right=20] node[midway, above] {$r^\prime$} (h);
        \end{tikzpicture}
    }
    \hfill
    \subfloat[][Composition\label{fig:composition}]{
        \centering
        \begin{tikzpicture}[node distance=0.7cm and 1.4cm, auto,
                main node/.style={circle, draw, minimum size=0.65cm, fill=blue!15},
                label node/.style={fill=white},
                line/.style={-Latex, line width=0.4mm}]
            \node[main node] (h) {h};
            \node[main node, right=of h] (n) {n};
            \node[main node, right=of n] (t) {t};

            \draw[line, dashed] (h) -- (n) node[midway, below] {$r_1$};
            \draw[line, dashed] (n) -- (t) node[midway, below] {$r_2$};
            \draw[line] (h) to[bend left=30] node[midway] {$r$} (t);
        \end{tikzpicture}
    }

    \caption{The four primary edge topological patterns we consider. We show \( (h, r, t) \) as the base triple common to all patterns, while the dashed lines with relations \(r^\prime\), \(r_1\) or \(r_2\), are edges that realize the defining feature of the pattern. We use \(r^\prime\), \(r_1\) or \(r_2\) to denote relations distinct from \(r\).}\label{fig:relational_patterns}
\end{figure}

We also define the following \textit{edge topological patterns} for triples \( (h,r,t) \in \mathcal{G} \):
\begin{enumerate}[a)]
    \item \( (h,r,t) \) \textit{is symmetric} \( \iff h \neq t \) and \( (t,r,h) \in \mathcal{G} \);
    \item \( (h,r,t) \) \textit{has inference} \( \iff \exists r'\neq r : (h,r',t) \in \mathcal{G} \);
    \item \( (h,r,t) \)  \textit{has inverse} \( \iff \exists r'\neq r : (t,r',h) \in \mathcal{G} \);
    \item \( (h,r,t) \)  \textit{has composition} \( \iff \exists r_1, r_2 \) and \( n \notin \{h,t\}: (h,r_1,n), (n, r_2, t) \in \mathcal{G} \).
\end{enumerate}

These edge topological patterns are illustrated in \cref{fig:relational_patterns}. It is important to note that a triple can satisfy multiple topological patterns simultaneously, or none at all.

While the above cardinalities and topological patterns are well defined for individual triples, previous studies~\citep{jin2023comprehensive,teneva2023_knowledge,bordes2013_transe,sun2019_rotate} have considered them as properties of relation types as a whole (e.g., relation \( r \) is said to be symmetric if \( \: \forall h,t, \;  \displaystyle{(h,r,t) \Rightarrow (t,r,h)}\)), despite the fact that a given property might be satisfied only by a fraction of the triples of a certain relation. We find that such aggregations lack proper formalization in the literature, with inconsistencies in how it is performed, and is prone to introducing noise in results (see \nameref{sec:results_overall}). Thus, we consider all the above defined properties at the individual triple level.

\subsection{Topological Patterns in Biomedical KGs}

To demonstrate how topological patterns can be observed in biomedical KGs, we consider examples from Hetionet.

\textbf{Symmetric edges:} Hetionet contains only a small number of symmetric edges (see \cref{tab:topology_patterns}), with all examples being of the \emph{Gene-regulates-Gene} relation type. In this instance, the symmetric edges represent that knockdown/overexpression of one gene significantly dysregulated the other~\cite{himmelstein2017_hetionet}. The creation of symmetric edges when the underlying data is not inherently directional (e.g., physical binding interactions between proteins) is a composition design decision that can, depending upon the scoring function chosen, be used when a KGE model shouldn't learn anything about the directionality of the edge.

\textbf{Inference edges:} Hetionet has several relation types that contain inference edges, for example for Gene-Disease links, where the \emph{Disease-upregulates-Gene} or \emph{Disease-downregulates-Gene} relation types are inference edges with \emph{Disease-associates-Gene}. Inference edges appear when there are logically or semantically distinct relationships between the same pair of entities. They can also arise when the relationship is captured in more than one underlying data source. Here, there is logically an association between a given disease and gene if an impact on gene expression has been measured. The relation types are also drawn from different sources (\emph{associates} from sources like DisGeNet and GWAS Catalog, \emph{up/downregulates} from a differential expression analysis~\cite{himmelstein2017_hetionet}).

\textbf{Inverse edges:} Inverse edges are rare in Hetionet, with all examples arising from combinations of different Gene-Gene relation types (a pair of genes being linked in one direction by \emph{regulation edge} and in the other by an \emph{interaction} edge). These examples may not be intentionally inverse and could be a product of the data integration process. Hypothetical inverse examples (\emph{Drug-treats-Disease} and \emph{Disease-treatedby-Drug} for example) could be added to a KG to enable simpler querying by an end user however.

\begin{figure}[h]
    \centering
    \includegraphics[width=\linewidth]{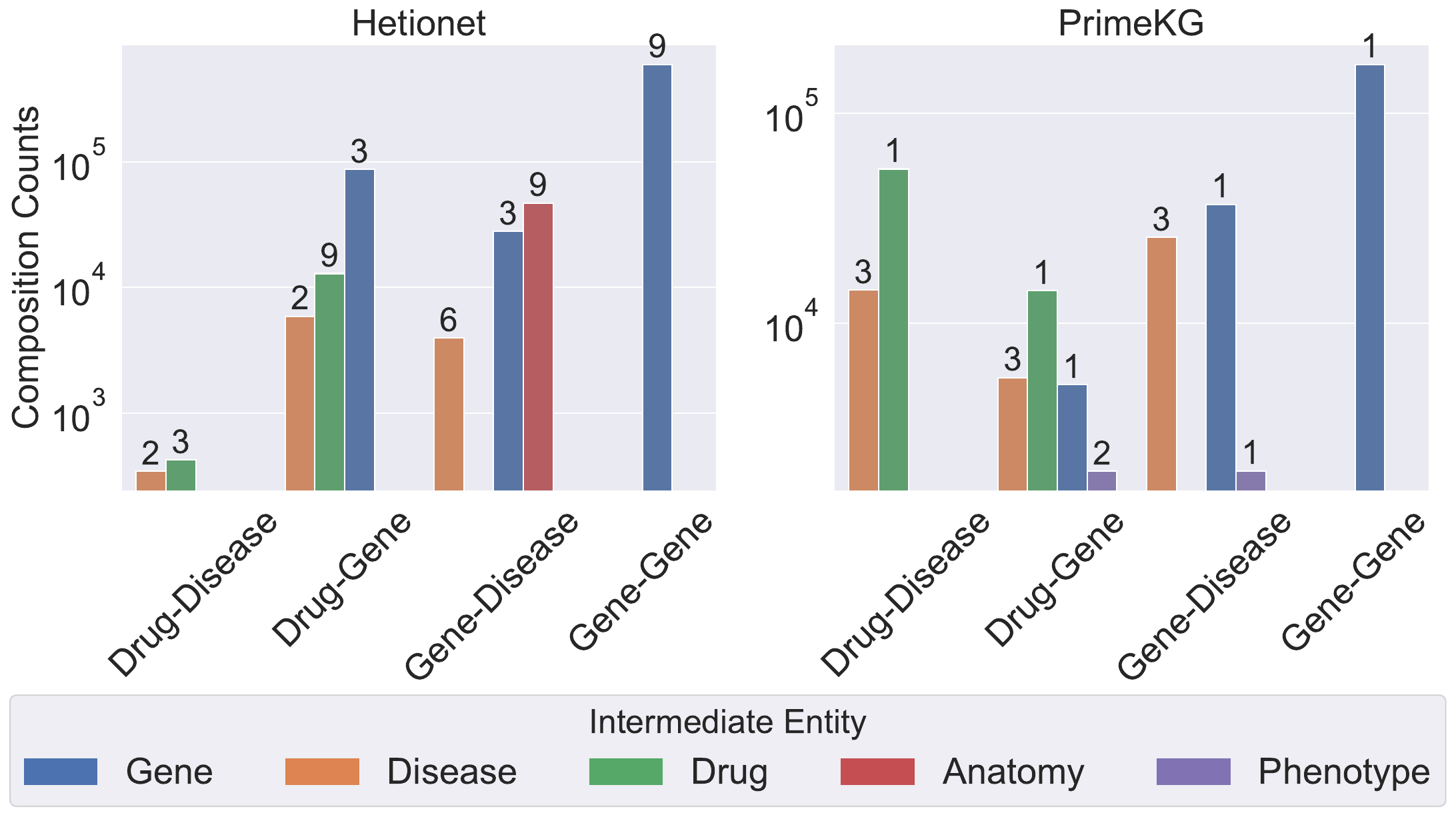}
    \caption{Counts of composition patterns for key relation types within Hetionet and PrimeKG. Values above the bars indicate the number of distinct pairs of relation types in the composition.}\label{fig:exemplar_patterns}
\end{figure}

\textbf{Composition edges:} Hetionet contains many relation types that are part of composition patterns (for example \emph{Compound-binds-Gene} being part of the composition pattern \emph{Compound-treats-Disease} and \emph{Disease-associates-Gene}), as they naturally arise in biomedical data. To illustrate this further, Figure~\ref{fig:exemplar_patterns} shows the counts of composition patterns connecting pairs of specific entity types, where we include PrimeKG in addition to Hetionet for comparison. These counts have been stratified for each edge by the type of the intermediate entity \( n \) in the composition pattern, and we show the number of unique patterns containing that entity type. This figure shows the abundance and diversity of this pattern across both datasets, where the most abundant composition edge is made up of three genes.

While all these four topological patterns are present in Hetionet, it is worth noting that some of the patterns are inherent to the nature of Biomedical KGs, while others may arise due to construction decisions taken during the creation of the KG.
\section{Experimental Setup}\label{sec:experimental_setup}

We investigate five public biomedical KGs: Hetionet \citep{himmelstein2017_hetionet}, PrimeKG\footnote{In the original PrimeKG graph, for every triple \((h,r,t)\) the reverse triple \((t,r,h)\) is present as well. We prepocessed PrimeKG to remove these reverse edges.} \citep{chandak2023_primekg}, PharmKG \citep{zheng2021_pharmkg}, OpenBioLink \citep{breit2020_openbiolink}, PharMeBINet \citep{konigs2022_pharmebinet} (detailed in \cref{tab:datasets}). These datasets are among the most common biomedical KGs used for benchmarking in the literature \citep{teneva2023_knowledge} and were chosen with the aim of encompassing a wide variety of sizes and topological patterns.
We also include the trivia KG FB15k-237 \citep{toutanova2015_fb15k-237} as a baseline, to detect any results unique to the biomedical domain.
On all KGs we train five of the most popular KGE models: TransE \citep{bordes2013_transe}, DistMult \citep{yang2015_embedding}, RotatE \citep{sun2019_rotate}, TripleRE \citep{yu2022_triplere} and ConvE \citep{detmers2018_conve}. Note that ConvE -- in addition to entity and relation embeddings -- also learns the parameters of a convolutional network, used in the scoring function (details in \cref{tab:scoring functions}). Whilst being all well-established among practitioners \citep{hu2020_ogb}, these models differ in complexity and in the relational patterns they can capture (\cref{tab:scoring functions}).
In particular, TransE and DistMult model the interaction between head entity, relation type and tail entity in quite simple terms. Nevertheless, despite missing certain topological patterns, they remain strong baselines. On the other hand, RotatE, TripleRE and ConvE are more recent approaches able to capture all four investigated topological patterns. Notice however that the theoretical capability of a scoring function to model a particular edge topological pattern is not per se a guarantee of stronger predictive performance on such edges \citep{jin2023comprehensive}. Our experiments are designed to quantify this impact.

The training scheme and hyperparameter optimisation details are presented in \cref{appendix:hyperparameter}.
The results reported in the following sections refer to tail predictions generated on the held out test split, by scoring each \((h,r,?)\) query against all entities in the KG and computing the rank of the ground truth tail \(t\), after masking out scores of other \((h,r,t')\) triples contained in the graph.
\section{Results}\label{sec:results}

\subsection{Topological Properties of Biomedical KGs}\label{sec:dataset_stats}

\begin{figure}[htb]
    \centering
    \includegraphics[width=.45\textwidth]{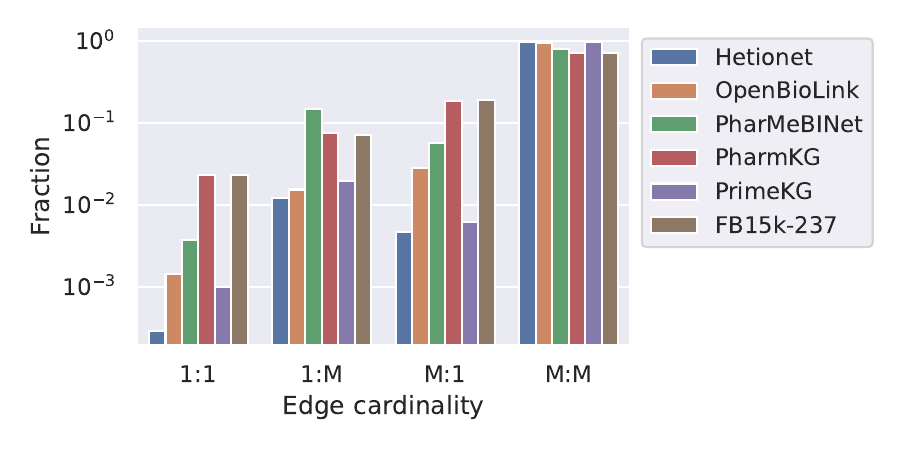}
    \caption{Occurrence of edge cardinalities in the datasets.}\label{fig:edge_cardinalities}
\end{figure}

\begin{figure}[htb]
    \centering
    \includegraphics[width=\linewidth]{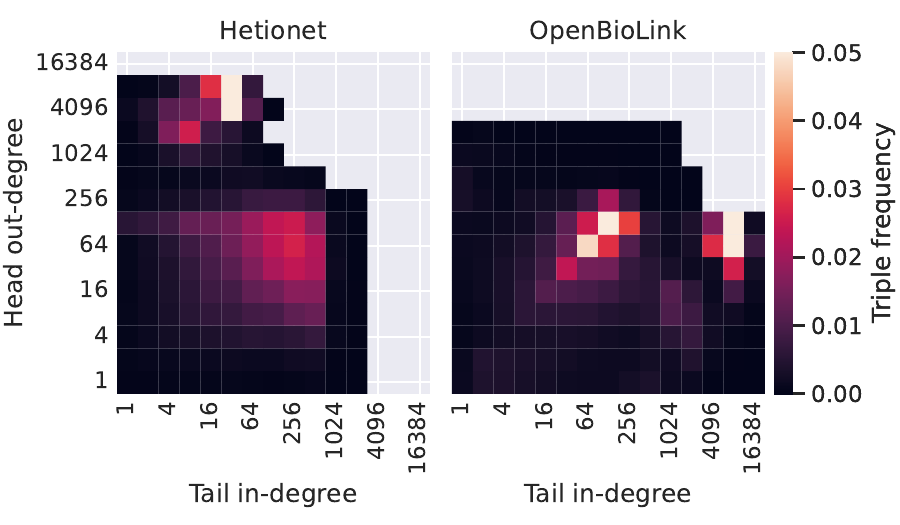}
    \caption{Relative frequency of triples when grouped by head out-degree and tail in-degree of the same relation type.}\label{fig:count_vs_degree}
\end{figure}

We first investigate how prevalent the different topological patterns are in the range of public biomedical KGs we consider.
Analysing this, we find that many-to-many is the predominant edge cardinality in all considered KGs, with only scarce occurrence of one-to-one, one-to-many, and many-to-one triples (\cref{fig:edge_cardinalities}).
This is in line with the majority of triples having an head out-degree and tail in-degree larger than one (\cref{fig:count_vs_degree} and \cref{fig:Appendix_count_vs_degree}).

Furthermore, we observe that edge topological patters vary in frequency across the different KGs (\cref{tab:topology_patterns}).
While \textit{composition} is the most prevalent pattern in all examined KGs, other edge topological patters occur at a high frequency in only a subset of KGs.
Of note is the lack of inverse edges in all the biomedical KGs, apart from OpenBioLink, highlighting that biomedical knowledge is typically encoded in a directed manner in these graphs

\begin{table}[hbt]
    \centering
    \caption{The fraction of edges with a given topological pattern (symmetry, inference, inverse, composition, or none of the four) in the datasets. As a single edge can satisfy multiple patterns, the sum of the frequencies in each row can exceed 1. Note that, since PrimeKG was pre-processed to remove reverse edges as described in \nameref{sec:experimental_setup}, it does not contain any symmetric triples.} \label{tab:topology_patterns}
    \resizebox{\columnwidth}{!}{%
        \begin{tabular}{lllllll}
            \toprule
            \textbf{Graph} & \textbf{Sym}    & \textbf{Inf}    & \textbf{Inv} & \textbf{Comp} & \textbf{None} \\
            \midrule
            Hetionet       & $0.002$         & $0.124$         & $0.001$      & $0.693$       & $0.298$       \\
            OpenBioLink    & $0.317$         & $0.372$         & $0.359$      & $0.840$       & $0.12$        \\
            PharMeBINet    & $\num{2.42e-4}$ & $0.052$         & $0.002$      & $0.598$       & $0.396$       \\
            PharmKG        & $0.197$         & $0.124$         & $0.059$      & $0.651$       & $0.279$       \\
            PrimeKG        & $0$             & $\num{2.08e-4}$ & $0$          & $0.807$       & $0.193$       \\
            \midrule
            FB15k-237      & $0.113$         & $0.161$         & $0.217$      & $0.645$       & $0.287$       \\
            \bottomrule
        \end{tabular}}
\end{table}

\subsection{Effect of Topological Properties on Model Accuracy}\label{sec:results_overall}

\begin{figure}[htb]
    \centering
    \includegraphics[width=0.95\columnwidth]{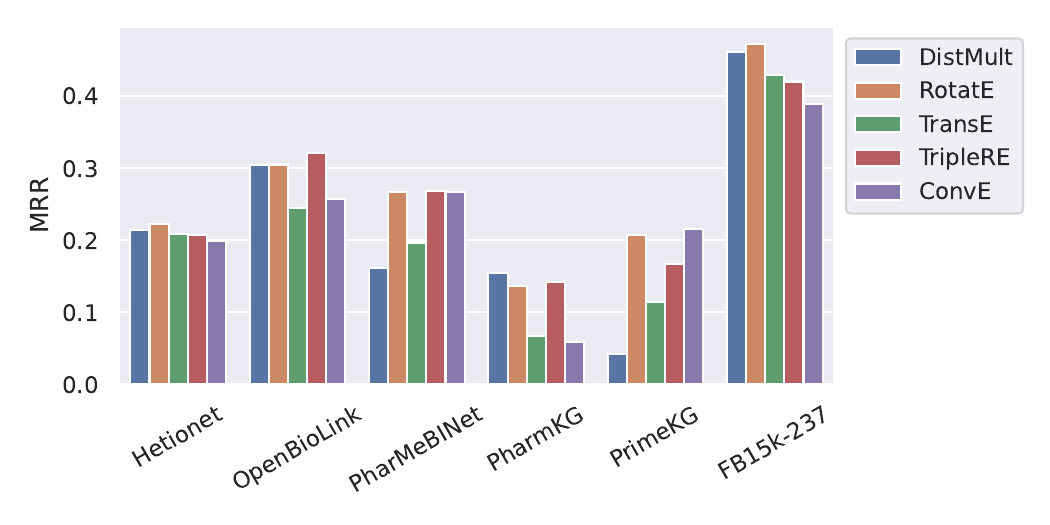}
    \caption{Mean reciprocal rank on the test split achieved by different KGE models, for the six datasets.}\label{fig:MRR}
\end{figure}

Overall, we observe a significant variance in mean reciprocal rank (MRR) achieved between the different KGs as well as between KGE models (\cref{fig:MRR}; see also \cref{fig:Appendix_hits_at_k} for Hits@1 and Hits@10, where we observe similar variability and consistent trends in the results).
In an attempt to understand the root cause of this variance, we focus our analysis on the differences in edge cardinalities and topological patterns.
When only considering the average occurrence of these topological properties per dataset, the data supports no conclusion about the effect on model accuracy of either edge cardinalities (\cref{fig:Appendix_avg_cardinalities_vs_MRR}) or edge topological patterns (\cref{fig:Appendix_avg_metrics_vs_MRR}).
Previous work (e.g., \citep{teneva2023_knowledge}) went one step further, classifying relation types based on the predominant edge topological pattern and cardinality and evaluating the model accuracy per relation type. However, this also does not result in a clear connection between topological properties and KGE model accuracy (\cref{fig:Appendix_avg_rel_properties_vs_MRR}).
We hypothesize that this could be due to the confounding effects of covariates, such as node degrees and different topological patterns, that are too difficult to disentangle, as these properties are often not homogenous within a given relation type.
Therefore, unlike prior work, we dissect the link between topological properties and model accuracy at the level of individual triples to allow for a finer-grained analysis with an improved statistical power and a reduced impact of covariates.

\begin{figure}[htb]
    \centering
    \includegraphics[width=.95\columnwidth]{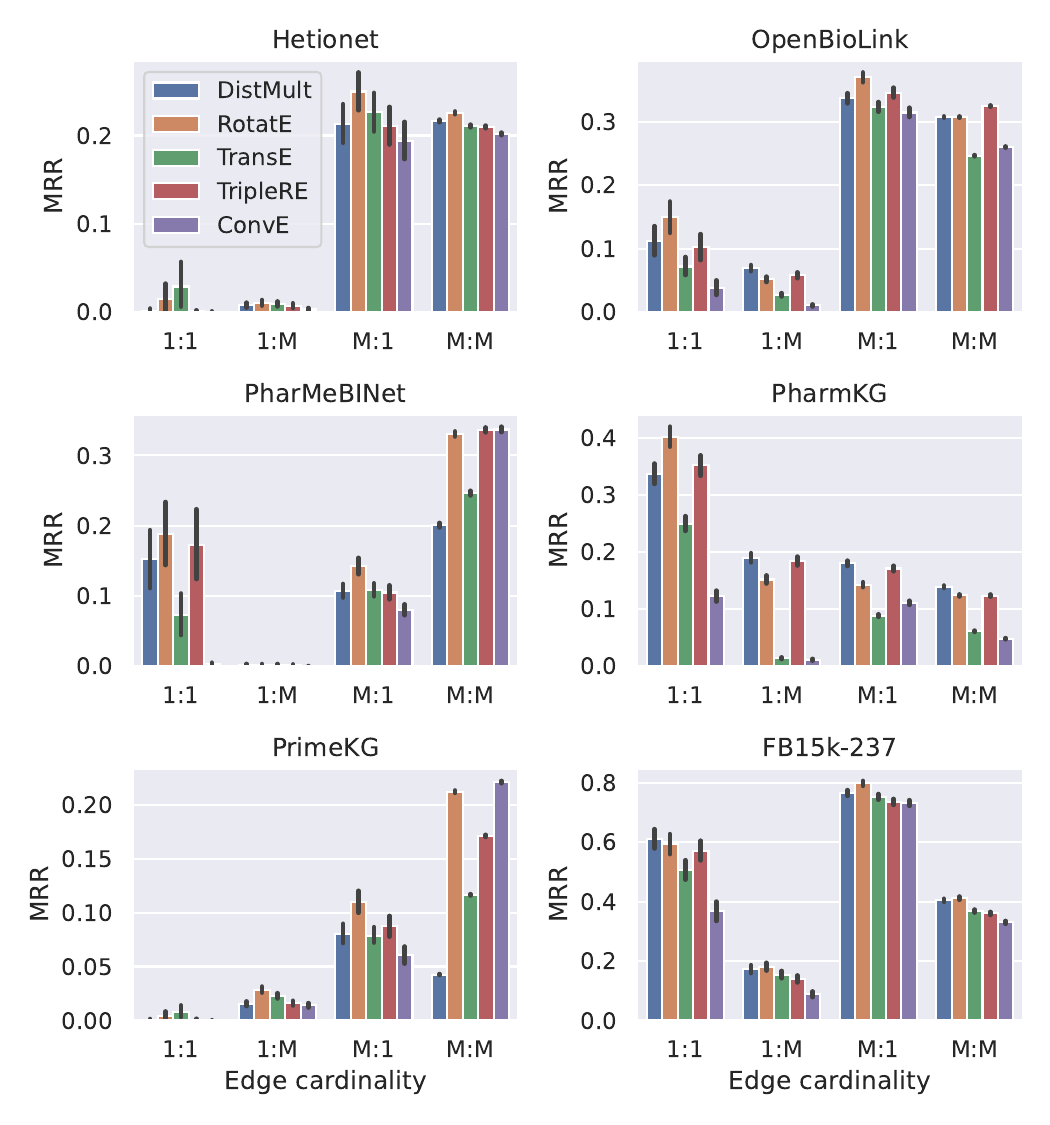}
    \caption{Effect of edge cardinality on MRR.}\label{fig:MRR_vs_cardinality}
\end{figure}

When comparing the edge cardinality of individual triples to the MRR achieved by the models on these triples, the observed effect strongly differs between the investigated KGs (\cref{fig:MRR_vs_cardinality}, and similarly for Hits@10 in \cref{fig:Appendix_Hits_at_10_vs_cardinality}). This result suggests that the exact entity degrees, together with other topological properties, might be better suited to explain the model accuracy than a binary one/many cardinality classification.

\begin{figure}[htb]
    \centering
    \subfloat{\includegraphics[width=0.48\linewidth]{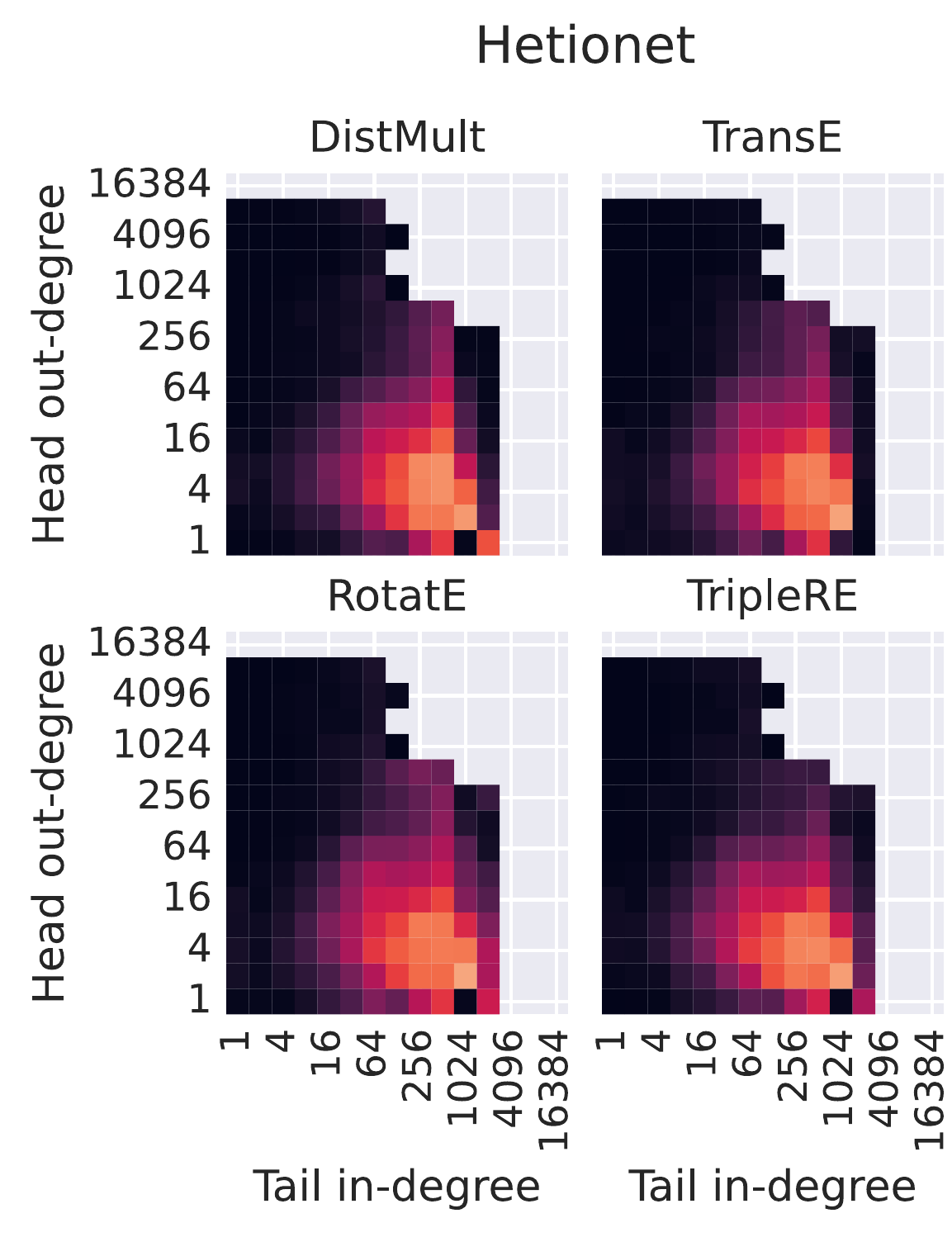}}
    \hspace{0.02\linewidth}
    \subfloat{\includegraphics[width=0.48\linewidth]{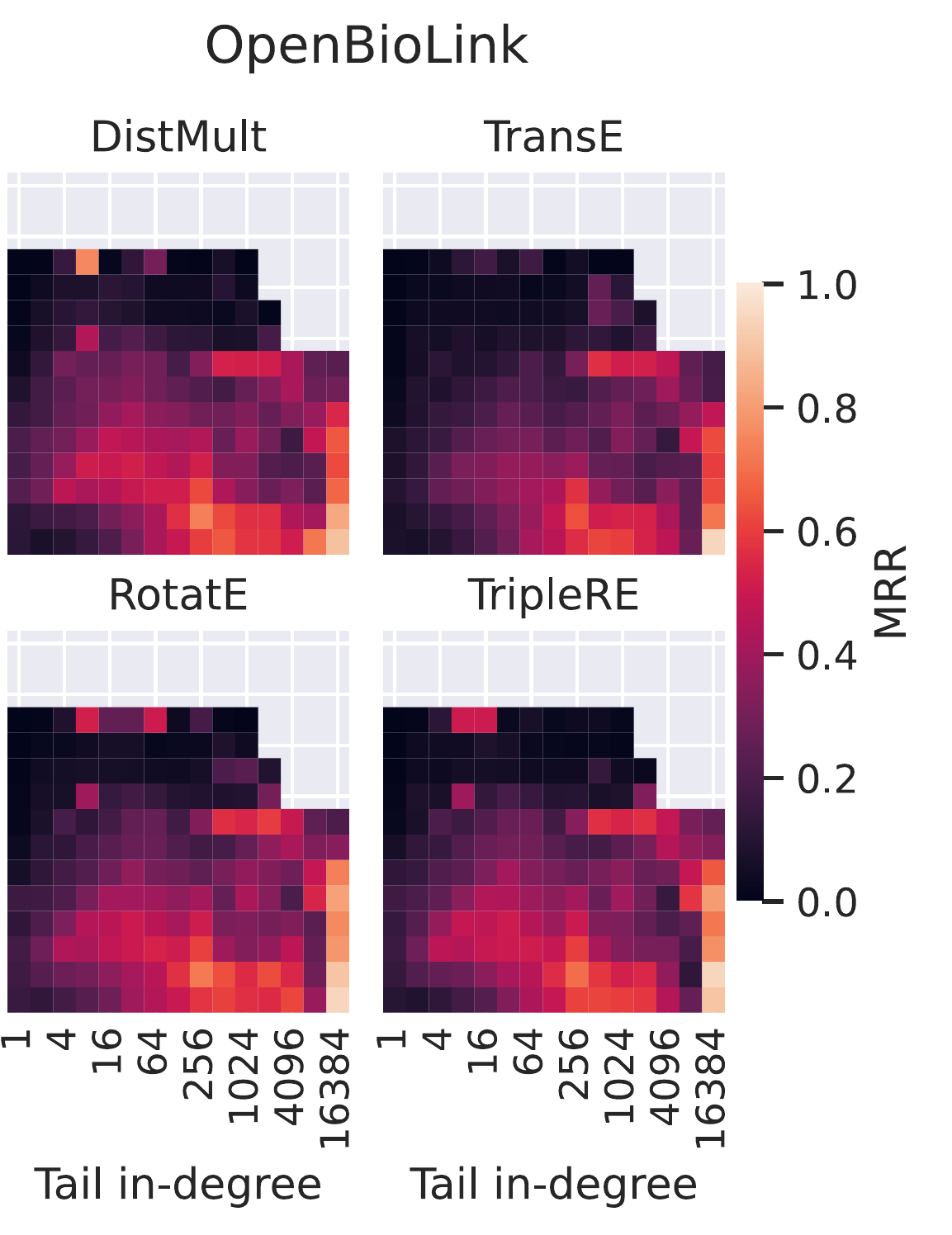}}
    \caption{Effect of the head and tail degrees of same relation type on MRR.}\label{fig:MRR_vs_degree}
\end{figure}

Indeed, for a given triple, we observe a strong positive correlation between model accuracy and the in-degree of the tail entity as well as a strong negative correlation between model accuracy and the out-degree of the head entity (\cref{fig:Appendix_MRR_degree_correlation}).
This is observed even more clearly when considering the degrees of same relation type (\cref{fig:MRR_vs_degree} and \cref{fig:Appendix_MRR_vs_degree,fig:Appendix_conve_MRR_vs_degree}; the same patterns are observed for Hits@10, as exemplified in \cref{fig:Appendix_hits_at_10_vs_degree}).
In fact, a high in-degree of the tail node in a tail prediction task has been linked to a higher score \citep{bonner2022_implications}, therefore increasing the likelihood of the model predicting it, as confirmed in \cref{fig:Appendix_rel_bias}.
On the other hand, a high out-degree of same relation type of the head node in a query \((h,r,?)\), implies multiple correct tail entities. Due to the incompleteness of the KG, some of these correct triples might not be present in the graph and thus cannot be filtered out during inference, making the task of predicting the specific entity that is expected by the model more challenging.

\begin{figure}[htb]
    \centering
    \subfloat{\includegraphics[width=0.48\linewidth]{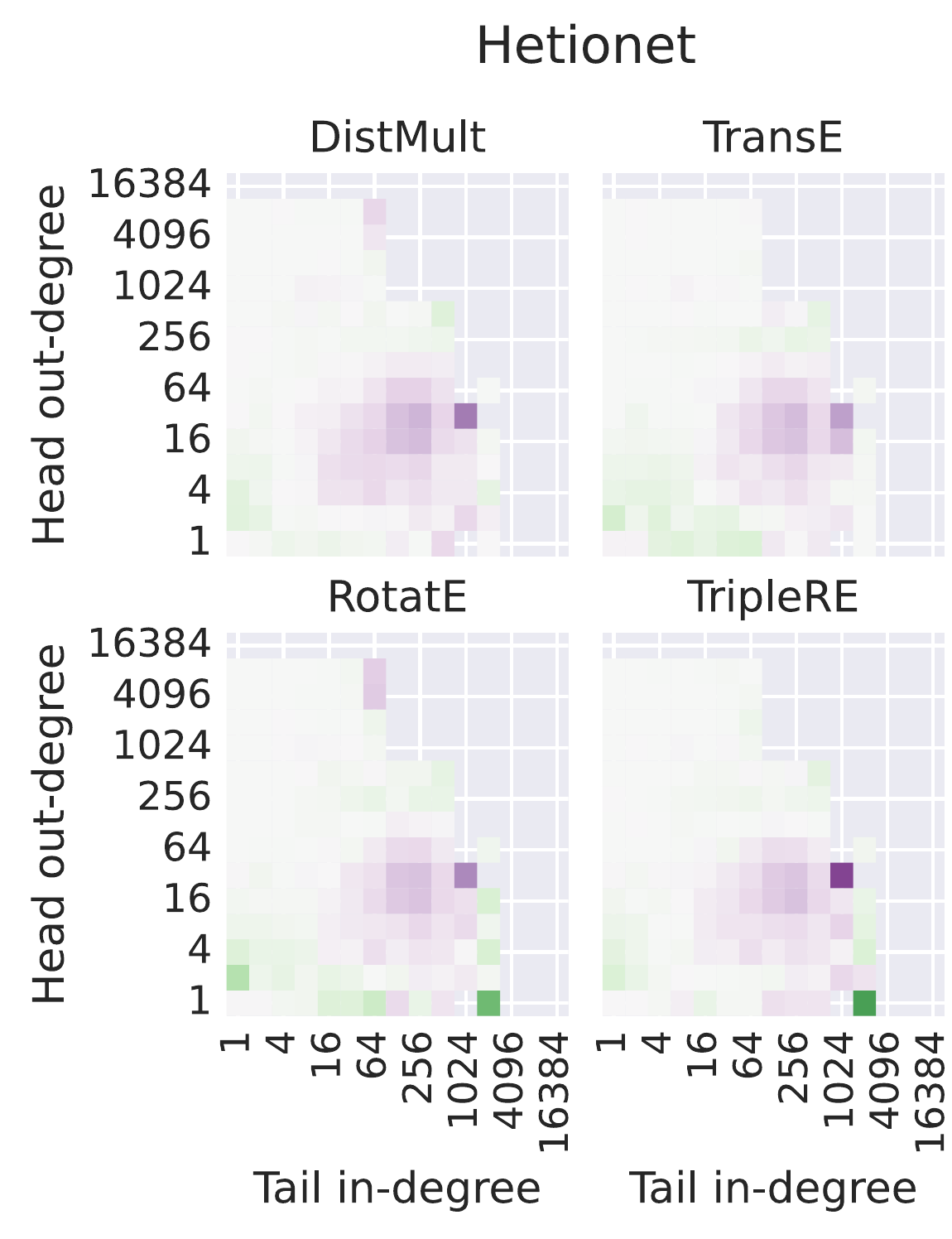}}
    \hspace{0.02\linewidth}
    \subfloat{\includegraphics[width=0.48\linewidth]{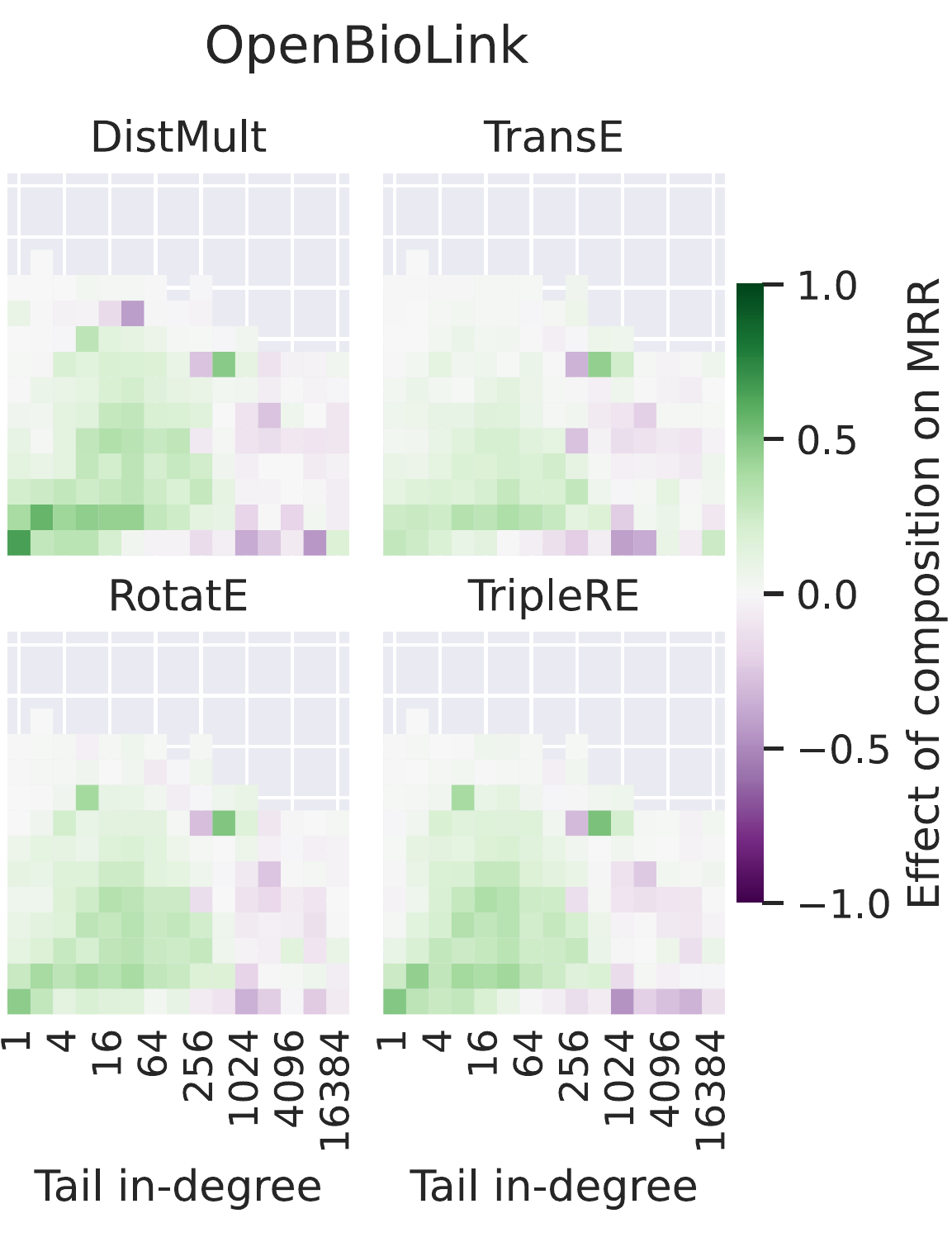}}
    \caption{The effect of having compositions on MRR, triples grouped by their head and tail degrees of same relation type.}\label{fig:MRR_vs_composition}
\end{figure}

Consequently, in an attempt to adjust for entity degree, we investigate the effect of edge topological patterns within sets of triples binned by head and tail degrees.
Overall, we find that the impact of edge topological patterns on model performance is more relevant if the degrees of head and tail entities are small.
When this is the case, compositions are beneficial to model accuracy across all datasets and models (see \cref{fig:MRR_vs_composition} and \cref{fig:Appendix_MRR_vs_composition,fig:Appendix_conve_MRR_vs_composition} for MRR; the same holds for Hits@10, e.g., \cref{fig:Appendix_hits_at_10_vs_composition}).
Interestingly, this is true even for \mbox{DistMult} that can't explicitly model compositional patterns.

When analysing the effect of the patterns \textit{is symmetric}, \textit{has inference} and \textit{has inverse}, we need to distinguish whether the counterpart of the tested edge was present in the training split.
If so, the prediction typically becomes much easier due to the availability of additional information seen during training \citep{toutanova2015_fb15k-237}, as shown in \cref{fig:Appendix_MRR_vs_symmetry_inference_inverse}.
An exception to this rule is the prediction of symmetric triples with the scoring function TransE: since TransE by design cannot model symmetric relations (\cref{tab:scoring functions}), it performs poorly on symmetric triples independently of the counterpart being present in the training data or not.
This factor is likely contributing to the subpar accuracy of TransE on OpenBioLink and PharmKG, which have a large fraction of symmetric triples (\cref{fig:MRR,tab:topology_patterns}).
On the other hand, our experiments confirm an above average performance of DistMult (which models all relations as symmetric) on symmetric triples, see \cref{fig:Appendix_MRR_vs_symmetry_inference_inverse}.
In the more challenging case, where the counterpart edge is not present in the training data, symmetry shows a detrimental effect on model accuracy, whereas having inverse or inference has only little impact on accuracy (\cref{fig:Appendix_MRR_vs_symmetry_inference_inverse}).

Note that all the above remarks apply equally to biomedical KGs and the general-domain KG FB15k-237.

\subsection{Predicting Specific Relation Types} \label{sec:specific_relations}

\begin{figure}[htb]
    \centering
    \includegraphics[width=0.95\linewidth]{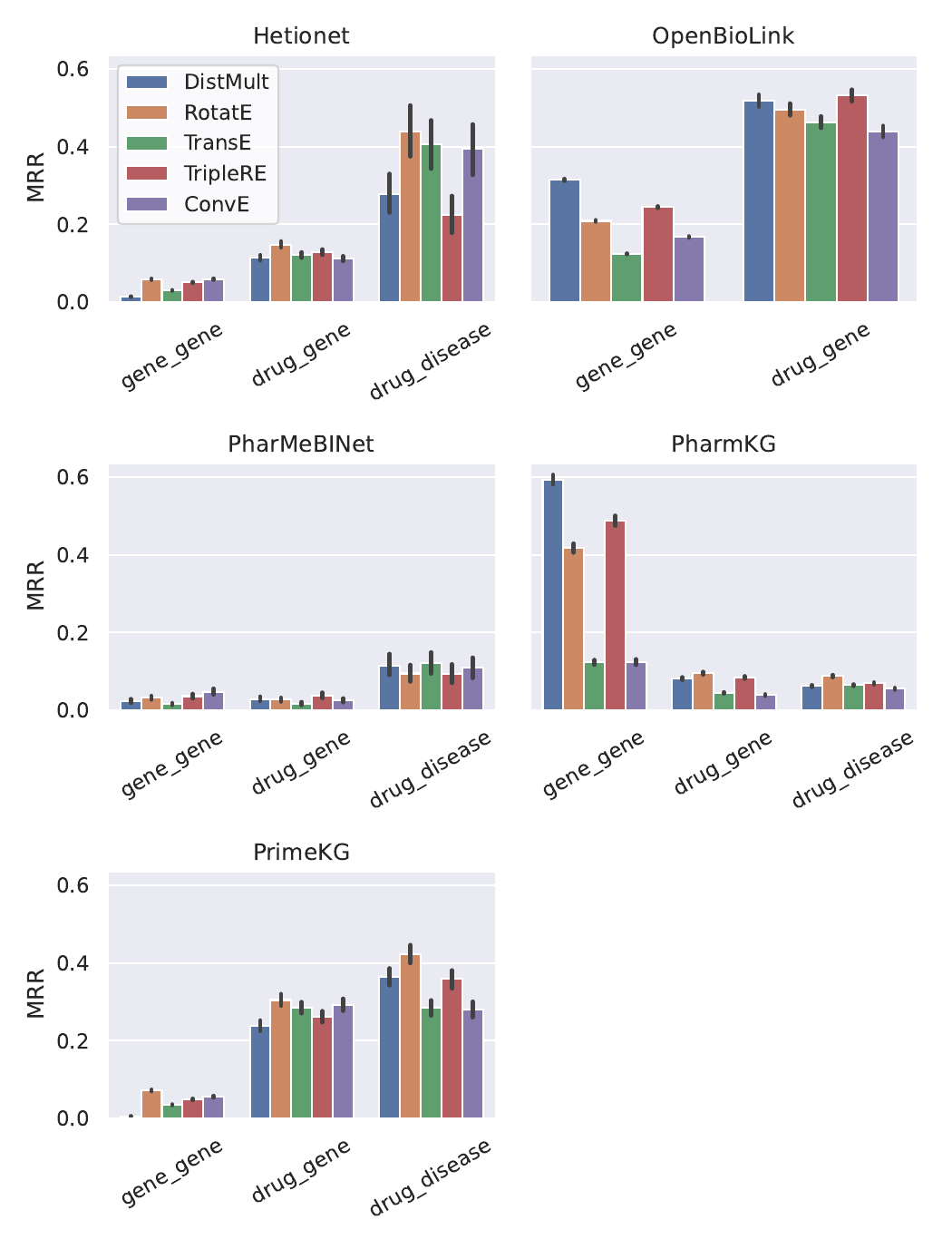}
    \caption{MRR of specific interaction types.}\label{fig:MRR_interesting_rels}
\end{figure}

To apply the analysis conducted in the previous sections to real-world tasks, we focus on the relations between pairs of entity types that are of particular interest to practitioners (Gene-Gene, Drug-Gene and Drug-Disease).
We investigate how they are represented in different biomedical KGs, and how these differences affect the predictive performance of KGE models. Statistics for the interactions of the considered entity types are given in \cref{fig:Appendix_interesting_relations_stats}, while model accuracy is plotted in \cref{fig:MRR_interesting_rels} (MRR) and \cref{fig:Appendix_hits_at_10_interesting_rels} (Hits@10).

One basic way of characterising the performance of a KGE model is to consider if it can rank entities of the correct type above those of other types when predicting a specific relation (for instance, gene entities when predicting drug-gene interactions) -- a task we refer to as \textit{demixing}. As displayed in \cref{fig:Appendix_demixing}, the trained models tend to consistently predict entities of the correct type, albeit with different levels of accuracy. Interestingly, DistMult often shows worse demixing capabilities than other scoring functions. As a consequence of the generally strong demixing performance, the size of the potential set of tails for a given interaction needs to be taken into consideration when comparing model accuracy across different datasets and relation types (the smaller the set of candidates, the higher the expected MRR achieved by random ranking).

\textbf{Gene-Gene}. As shown in \cref{fig:interesting_relations_stats_a}, these interactions are characterized by a large number of potential tails and a large head out-degree of same relation type (with the exception of PharmKG for the latter), making predictions hard. However, a high predictive accuracy is observed on PharmKG and OpenBioLink: this is explained by the fact that, for these datasets only (\cref{fig:interesting_relations_stats_b}), most triples are symmetric (and a significant portion also have inverse/inference edges), with the counterpart likely to have been seen during training. This also explains the relative ordering of scoring functions for OpenBioLink and PharmKG, with DistMult being the best and TransE the worst. Interestingly, ConvE also struggles on these highly symmetric datasets.

\textbf{Drug-Gene}. In Hetionet, PrimeKG and OpenBioLink these, interactions are easier to predict for the models than gene-gene interactions (\cref{fig:MRR_interesting_rels}). This is likely due to an overall smaller number of potential tails seen during training and, in the case of PrimeKG, a markedly smaller head out-degree. PharmKG and PharMeBINet do not satisfy either of these properties, and indeed we observe no improvements in MRR/Hits@10. The strong predictive power observed for OpenBioLink can be linked to the presence of inverse edges for the vast majority of these triples.

\textbf{Drug-Disease}. Despite a small number of training triples (which is reflected in the sub-optimal demixing profile, \cref{fig:Appendix_demixing_dd}), drug-disease interactions are easily predicted in Hetionet as only 91 disease entities appear as tails of such triples. Remarkably good predictive performance across all scoring functions is observed for PrimeKG, where the number of candidate tails is larger but the out-degree of head entities remains contained. When both these parameters increase (as in PharMeBINet and PharmKG) performance visibly degrades, despite larger in-degree of tails.

\subsection{Case Study: Effect of Additional Training Data}\label{subsec:hetio_phbnet_comparison}

Even when focussing on specific interactions that are represented in multiple KGs, a direct comparison of KGE predictive power across different datasets is limited by having to compare different test sets. In the case of Hetionet and PharMeBINet, however, the latter is constructed by augmenting the former with data coming from other biological databases~\citep{konigs2022_pharmebinet}. One can therefore often connect individual triples in Hetionet to their exact match in the larger PharMeBINet, which opens up the possibility of studying at a more fundamental level the impact of additional training data on the predictive performance of KGE models. This is important to practitioners as they construct the training KG for a specific task, which is usually done by sub-sampling relevant triples from larger (often proprietary) databases~\cite{chandak2023_primekg}.

\begin{figure}[htb]
    \centering
    \includegraphics[width=0.95\linewidth]{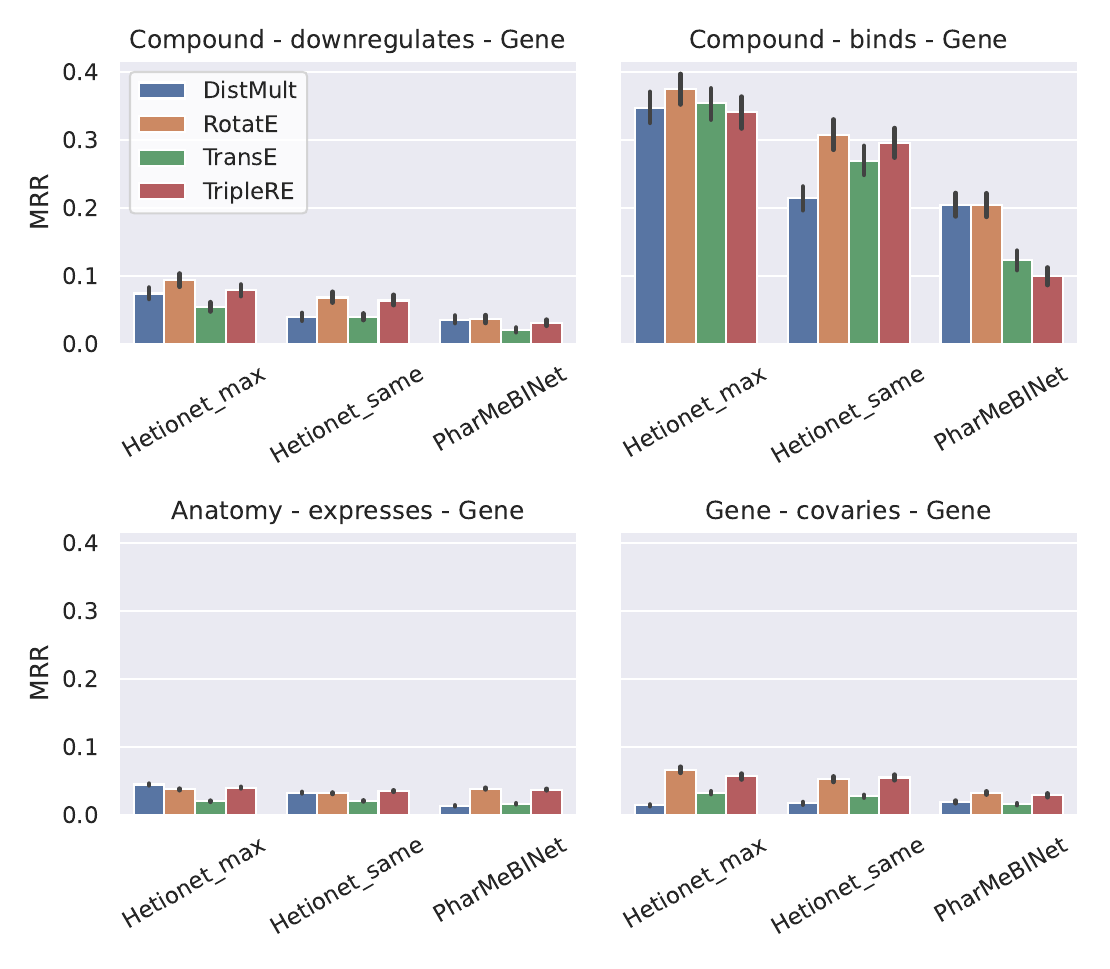}
    \caption{Comparison of models trained on Hetionet and PharMeBINet when testing MRR on a set of common edges.}\label{fig:hetio_phbnet_MRR}
\end{figure}

We consider eight relation types where we can find a significant number of common triples between the two datasets (statistics are given in \cref{tab:Appendix_hetio_phbnet_comparison}). We extract 10\% of the shared triples of the considered relation type as test set and use all other edges in each graph for training. This requires retraining each KGE model separately for each investigated relation type. Due to the higher computational complexity of ConvE and its resemblance of the other investigated scoring functions throughout all results, we restrict this investigation to the four shallow embedding models.
In the case of Hetionet, in addition to the embedding size maximizing memory utilization, we also train models with the same embedding size chosen for PharMeBINet, which is generally strictly smaller (\textit{Hetionet\_max} and \textit{Hetionet\_same} in \cref{fig:hetio_phbnet_MRR}; hyperparameters in \cref{tab:Appendix_hyperparams}). To further ensure a fair comparison, at test time we restrict predictions to a custom set of candidate tails that is the same for the two datasets, namely the set of entities appearing as tails in shared triples of the given relation type.

We only report on MRR, as (similarly to previous sections) Hits@10 shows analogous patterns. As displayed in \cref{fig:hetio_phbnet_MRR} and \cref{fig:Appendix_hetio_phbnet_other_rels}, across all tested relation types there is no indication that the KGE models are able to benefit from the additional data seen when training on the larger PharMeBINet graph. On the contrary, models trained on Hetionet consistently perform better, even when using the same embedding size. Interestingly, the gap in MRR varies strongly across relation types and also across scoring functions, with DistMult showing generally little difference when comparing models trained with the same embedding size, while distance-based scoring functions experience worse degradation. From \cref{tab:Appendix_hetio_phbnet_comparison} we notice that the relation types where the MRR gap is more significant tend to have a larger overall head out-degree (and more relation types coming out from head nodes) for triples in PharMeBINet, compared to Hetionet. We hypothesise that the fact that these nodes are used as head entities for more relation types and triples, many of which are likely not relevant for the specific task at hand, negatively impacts the quality of embeddings. On the other hand, \textit{Compound-downregulates-Gene} and \textit{Compound-causes-Side Effect} exemplify relation types where PharMeBINet contains far more triples in addition to the ones in Hetionet. Even in this case, where we could expect the additional training data in the larger dataset to be strictly relevant to the prediction task, all scoring functions show markedly degraded performance when trained on PharMeBINet.

We note that the intention of these results is not to state that one dataset is superior to the other, but rather to highlight any observed differences in how models are able to use training data.
However, these results do suggest that, in scenarios where the memory budget is fixed, training on smaller, tailored graphs and increasing the embedding size could be more beneficial than expanding the size of the KG, as the additional data can be a source of confusion for shallow KGE models.

\section{Conclusions}

This paper analyses the topological properties of widely used biomedical KGs and compares the corresponding predictive performance of different KGE models, focussing in particular on link-prediction tasks relevant to practitioners. Deviating from previous studies, we justify the need to look at properties of individual edges when trying to explain results, as pooling at the relation level introduces too much noise to detect any patterns. By going beyond the coarse-level one/many binary classification typically used for edge cardinality, we find that considering the actual degrees of head and tail nodes gives a stronger predictor explaining model accuracy. Interestingly, similar interpretations apply to both biomedical KGs and the general-domain KG.
Edge topological patterns also impact predictive accuracy, especially when entity degrees are small. For such patterns, we observe an improved accuracy when the counterpart edge (e.g., the reverse edge for symmetric triples) has been seen during training. This in turn raises the problem of inconsistent model validation schemes and topological property definitions when comparing results of previous studies on similar datasets. We address this problem by releasing all predictions from our experiments, together with a new toolbox for KG topological analysis.
Finally, by performing a case study comparing predictions on common sets of edges shared by different KGs, we show that training on larger graphs, encoding more biomedical data, can unintuitively harm predictive performance. This evidence should encourage a wider discussion on the guiding principles to adopt when constructing the training KG for biomedical tasks -- a crucial problem for real-world practitioners, but scarcely investigated in the literature.

\bibliographystyle{icml2021}
\bibliography{kge_paper}

\onecolumn
\appendix
\renewcommand\thefigure{\thesection.\arabic{figure}}
\setcounter{figure}{0}
\renewcommand\thetable{\thesection.\arabic{table}}
\setcounter{table}{0}

\renewcommand{\topfraction}{1.0}
\renewcommand{\bottomfraction}{1.0}
\renewcommand{\textfraction}{0.0}

\section{Dataset and Model Properties}\label{appendix:dataset_model}

\begin{table}[hbt]
    \centering
    \caption{Properties of the considered datasets.}\label{tab:datasets}
    \begin{tabular}{lllll}
        \toprule
        \textbf{Graph} & \textbf{\# Entities} & \textbf{\# Relations} & \textbf{\# Triples} & \textbf{Avg node degree} \\
        \midrule
        Hetionet       & 45,158               & 24                    & 2,250,197           & 99.66                    \\
        OpenBioLink    & 184,635              & 28                    & 4,563,405           & 49.43                    \\
        PharMeBINet    & 2,653,751            & 208                   & 15,883,653          & 11.97                    \\
        PharmKG        & 188,296              & 39                    & 1,093,236           & 11.61                    \\
        PrimeKG        & 129,375              & 30                    & 4,050,064           & 62.61                    \\
        \midrule
        FB15k-237      & 14,541               & 237                   & 310,116             & 42.65                    \\
        \bottomrule
    \end{tabular}
\end{table}

\begin{figure}[htb]
    \centering
    \includegraphics[width=\textwidth]{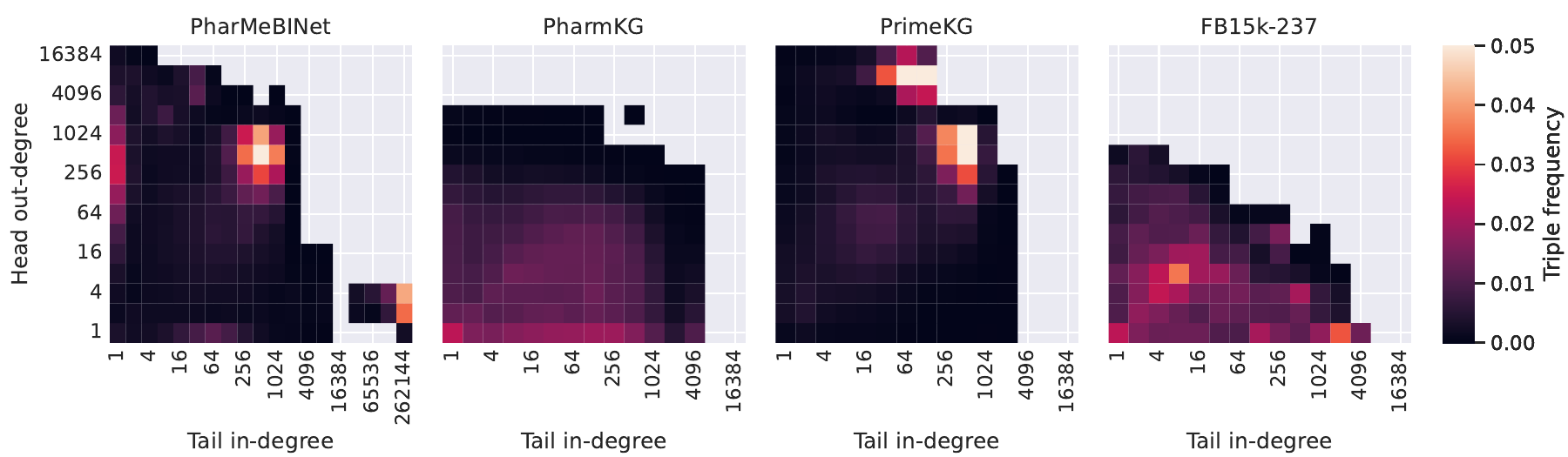}
    \caption{Relative frequency of triples when grouped by head out-degree and tail in-degree of the same relation type.} \label{fig:Appendix_count_vs_degree}
\end{figure}

\begin{table}[H]
    \centering
    \caption{Scoring functions and their ability to model four fundamental relation properties: S = Symmetry; INF = Inference; INV = Inversion; C = Composition. For RotatE we assume $d$ even and denote by $\C^{\frac{d}{2}}$ the vector space $\mathbb{R}^d = (\mathbb{R} \oplus i\mathbb{R})^{\frac{d}{2}}$ with the structure of $\R$-algebra induced by the product of complex numbers. $p \in \left\{ 1,2 \right\}$; $\circ$ denotes the Hadamard product, $*w$ is the application of a convolutional filter $w$.}\label{tab:scoring functions}
    \begin{tabular}{lllcccc}
        \toprule
        \textbf{Model} & \multicolumn{2}{c}{\textbf{Scoring function}}                & \textbf{S}                                                & \textbf{INF} & \textbf{INV} & \textbf{C}          \\
        \midrule
        TransE         & $-\norm{\vh + \vr - \vt}_p$                                  & $\vh, \vr, \vt \in \R^d$                                  & \xmark       & \cmark       & \cmark     & \cmark \\
        RotatE         & $-\norm{\vh \circ e^{i \vr} - \vt}_p$                        & $\vh, \vt \in \C^{\frac{d}{2}}, \vr \in \R^{\frac{d}{2}}$ & \cmark       & \cmark       & \cmark     & \cmark \\
        DistMult       & $\langle \vr, \vh, \vt \rangle$                              & $\vh, \vr, \vt \in \R^d$                                  & \cmark       & \cmark       & \xmark     & \xmark \\
        TripleRE       & $-\norm{\vh \circ \bm{r_h} + \bm{r_m} - \vt \circ \bm{r_t}}$ & $\vh, \bm{r_h}, \bm{r_m}, \vt, \bm{r_t} \in \R^d$         & \cmark       & \cmark       & \cmark     & \cmark \\
        ConvE          & $\langle \textrm{MLP}([\vh,\vr] * w) , \vt\rangle$           & $\vh, \vr, \vt \in \R^d, w \in \R^{3 \times 3}$           & \cmark       & \cmark       & \cmark     & \cmark \\
        \bottomrule
    \end{tabular}
\end{table}

\newpage
\FloatBarrier
\section{Details on Experimental Setup and Hyperparameter Selection}\label{appendix:hyperparameter}
\setcounter{figure}{0} 
\setcounter{table}{0} 

All datasets were randomly split into training, validation and test set ($80$\% / $10$\% / $10$\%; in the case of PharMeBINet, $99.3$\% / $0.35$\% / $0.35$\% to mitigate the increased inference cost on the larger dataset).
To ensure comparability across KGs, this random split was used even if pre-defined training, validation and test sets were provided with a dataset.
We adopted log-sigmoid loss with negative adversarial sampling \citep{sun2019_rotate} and margin $12.0$, and the Adam optimiser \citep{kingma2015_adam} for updating parameters. During training we always used negative sample sharing \cite{cattaneo2022_bess}. All experiments were performed on Graphcore IPUs using the BESS framework\footnote{\url{https://github.com/graphcore-research/bess-kge}} \citep{cattaneo2022_bess}.
A fixed batch size of $768$ triples per device ($192$ for PharMeBINet) was adopted, while the embedding size for entities and relations was chosen for each KG and each scoring function independently to maximise the memory utilisation of a Bow-2000 IPU machine with 4 IPU processors (in the case of PharMeBINet, a Bow Pod$_{16}$ with 16 IPUs).
This is to ensure a fair comparison between scoring functions with different memory costs. For some scoring functions, especially DistMult, the memory footprint is typically dominated by the model parameters, allowing a larger hidden size for smaller KGs. For other scoring functions, especially TripleRE and ConvE, memory is typically dominated by activations, resulting in a similar hidden size for differently sized KGs.
The learning rate, the norm used by the scoring function ($\mathrm{L}1$ or $\mathrm{L}2$) and the number of negative samples were determined by a hyperparameter sweep, based on the validation MRR (\cref{tab:Appendix_hyperparams}). The full configurations can be found in the training scripts, released with the code\footnote{\url{https://github.com/graphcore-research/kg-topology-toolbox/tree/main/the_role_of_graph_topology_paper/train}}.

\begin{table}[hbt]
    \centering
    \caption{Experiment hyperparameters for different datasets. \textit{Hetionet\_same} refers to the alternative experimental configuration used in \cref{subsec:hetio_phbnet_comparison}.}\label{tab:Appendix_hyperparams}
    \begin{tabular}{llllll}
        \toprule
        \textbf{Graph}                  & \textbf{Model} & \textbf{Hidden size} & \textbf{Learning Rate} & \textbf{Scoring Norm} & \textbf{\# Negative samples / positive} \\
        \midrule
        \multirow{4}{*}{Hetionet}       & DistMult       & $2048$               & $0.0003$               & -                     & $16$                                    \\
                                        & RotatE         & $512$                & $0.001$                & $\mathrm{L}2$         & $16$                                    \\
                                        & TransE         & $1024$               & $0.0001$               & $\mathrm{L}1$         & $16$                                    \\
                                        & TripleRE       & $384$                & $0.0001$               & $\mathrm{L}1$         & $16$                                    \\
                                        & ConvE          & $676$                & $0.001$                & -                     & $16$                                    \\
        \midrule
        \multirow{4}{*}{Hetionet\_same} & DistMult       & $300$                & $0.0003$               & -                     & $16$                                    \\
                                        & RotatE         & $128$                & $0.003$                & $\mathrm{L}2$         & $16$                                    \\
                                        & TransE         & $256$                & $0.0003$               & $\mathrm{L}1$         & $16$                                    \\
                                        & TripleRE       & $256$                & $0.0001$               & $\mathrm{L}1$         & $16$                                    \\
        \midrule
        \multirow{4}{*}{OpenBioLink}    & DistMult       & $768$                & $0.0003$               & -                     & $16$                                    \\
                                        & RotatE         & $256$                & $0.003$                & $\mathrm{L}2$         & $16$                                    \\
                                        & TransE         & $512$                & $0.0001$               & $\mathrm{L}1$         & $16$                                    \\
                                        & TripleRE       & $256$                & $0.001$                & $\mathrm{L}2$         & $16$                                    \\
                                        & ConvE          & $529$                & $0.001$                & -                     & $16$                                    \\
        \midrule
        \multirow{4}{*}{PharMeBINet}    & DistMult       & $300$                & $0.003$                & -                     & $16$                                    \\
                                        & RotatE         & $128$                & $0.001$                & $\mathrm{L}2$         & $16$                                    \\
                                        & TransE         & $256$                & $0.00003$              & $\mathrm{L}1$         & $16$                                    \\
                                        & TripleRE       & $256$                & $0.0001$               & $\mathrm{L}2$         & $16$                                    \\
                                        & ConvE          & $256$                & $0.0001$               & -                     & $16$                                    \\
        \midrule
        \multirow{4}{*}{PharmKG}        & DistMult       & $768$                & $0.003$                & -                     & $16$                                    \\
                                        & RotatE         & $384$                & $0.003$                & $\mathrm{L}2$         & $16$                                    \\
                                        & TransE         & $768$                & $0.0001$               & $\mathrm{L}1$         & $16$                                    \\
                                        & TripleRE       & $384$                & $0.0003$               & $\mathrm{L}1$         & $16$                                    \\
                                        & ConvE          & $529$                & $0.001$                & -                     & $16$                                    \\
        \midrule
        \multirow{4}{*}{PrimeKG}        & DistMult       & $1024$               & $0.0003$               & -                     & $16$                                    \\
                                        & RotatE         & $384$                & $0.001$                & $\mathrm{L}2$         & $16$                                    \\
                                        & TransE         & $768$                & $0.0001$               & $\mathrm{L}1$         & $16$                                    \\
                                        & TripleRE       & $256$                & $0.0001$               & $\mathrm{L}1$         & $16$                                    \\
                                        & ConvE          & $576$                & $0.0003$               & -                     & $16$                                    \\
        \midrule
        \multirow{4}{*}{FB15k-237}      & DistMult       & $4096$               & $0.001$                & -                     & $16$                                    \\
                                        & RotatE         & $1024$               & $0.003$                & $\mathrm{L}2$         & $16$                                    \\
                                        & TransE         & $2048$               & $0.0001$               & $\mathrm{L}1$         & $16$                                    \\
                                        & TripleRE       & $256$                & $0.001$                & $\mathrm{L}1$         & $16$                                    \\
                                        & ConvE          & $676$                & $0.001$                & -                     & $16$                                    \\
        \bottomrule
    \end{tabular}
\end{table}

\clearpage
\FloatBarrier
\section{Additional Results}
\setcounter{figure}{0} 
\setcounter{table}{0} 

\begin{figure}[htb]
    \centering
    \includegraphics[width=0.9\columnwidth]{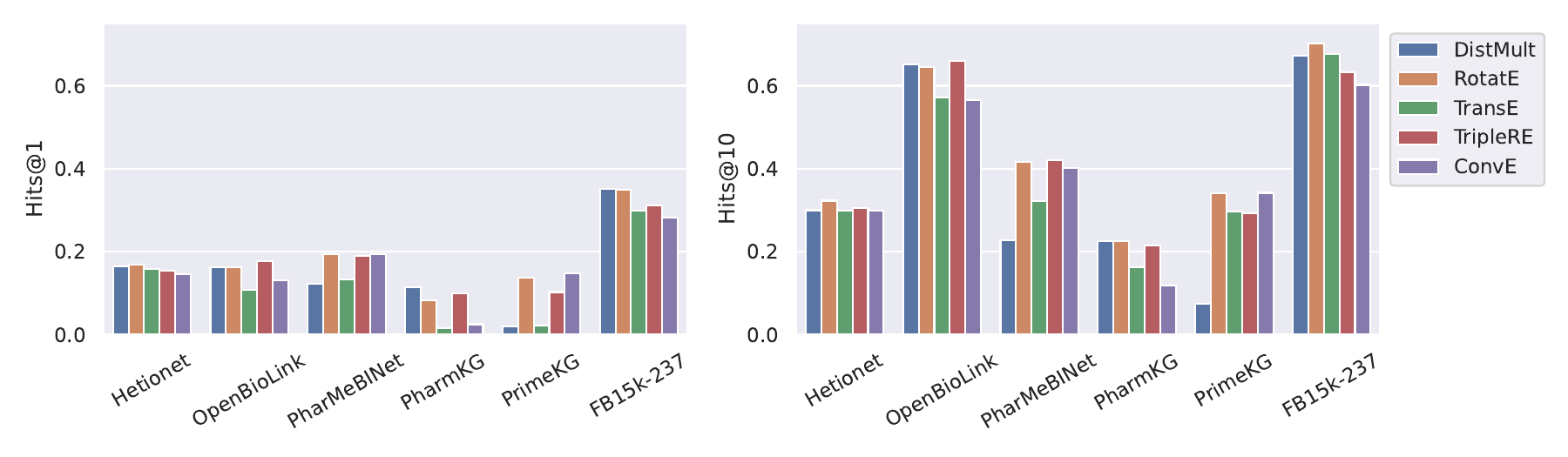}
    \caption{Hits@1 and Hits@10 on the test split achieved by the KGE models, for the six datasets.}\label{fig:Appendix_hits_at_k}
\end{figure}

\begin{figure}[htb]
    \centering
    \includegraphics[width=.9\columnwidth]{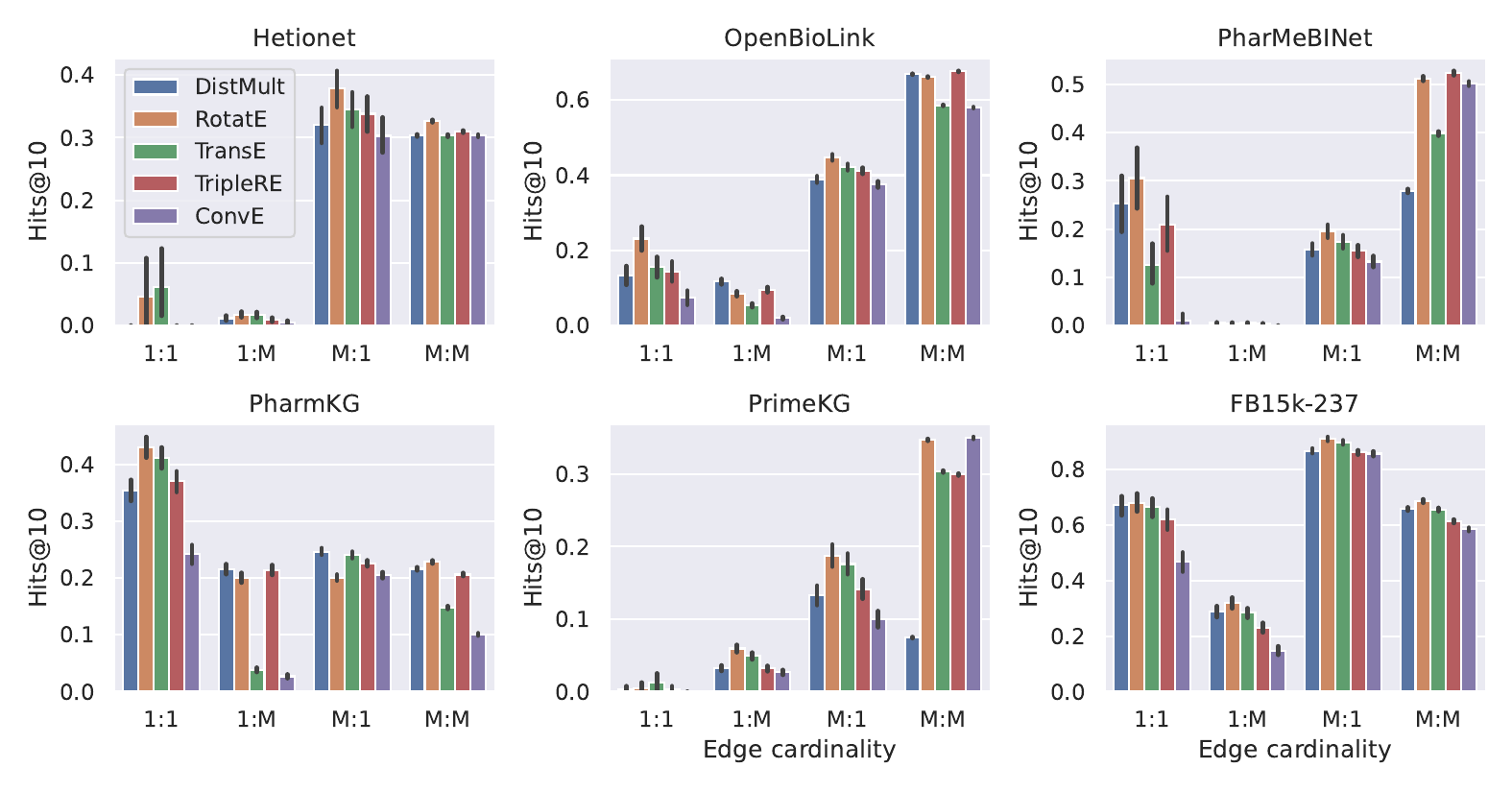}
    \caption{Effect of edge cardinality on Hits@10.}\label{fig:Appendix_Hits_at_10_vs_cardinality}
\end{figure}

\begin{figure}[htb]
    \centering
    \subfloat[][\label{fig:Appendix_avg_cardinalities_vs_MRR}]{
        \includegraphics[width=0.95\textwidth]{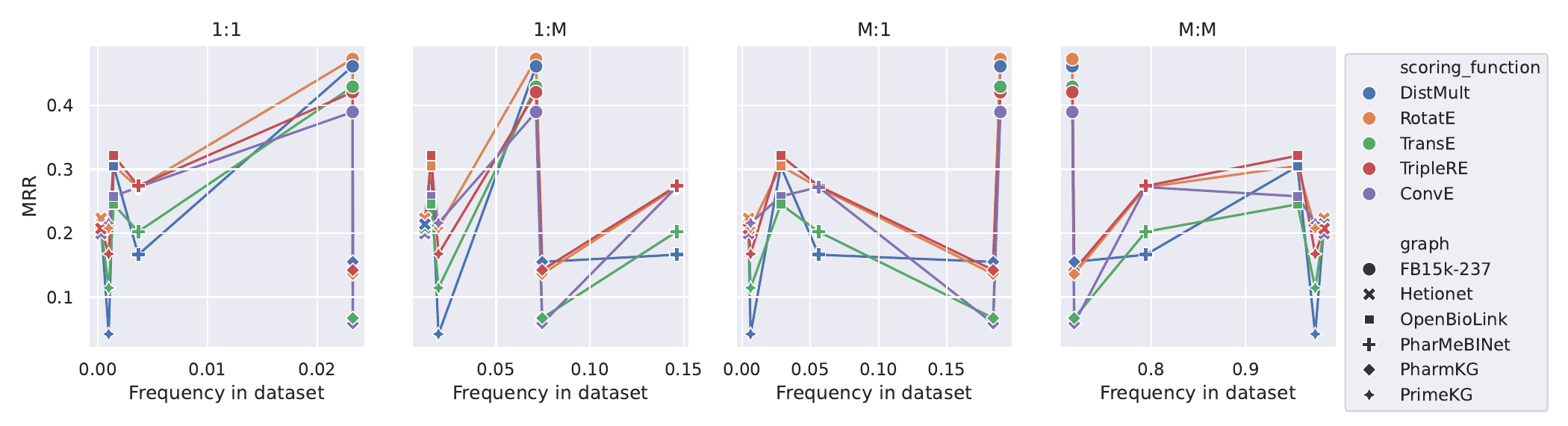}
    }\\
    \vspace{2pt}
    \subfloat[][\label{fig:Appendix_avg_metrics_vs_MRR}]{
        \includegraphics[width=0.95\textwidth]{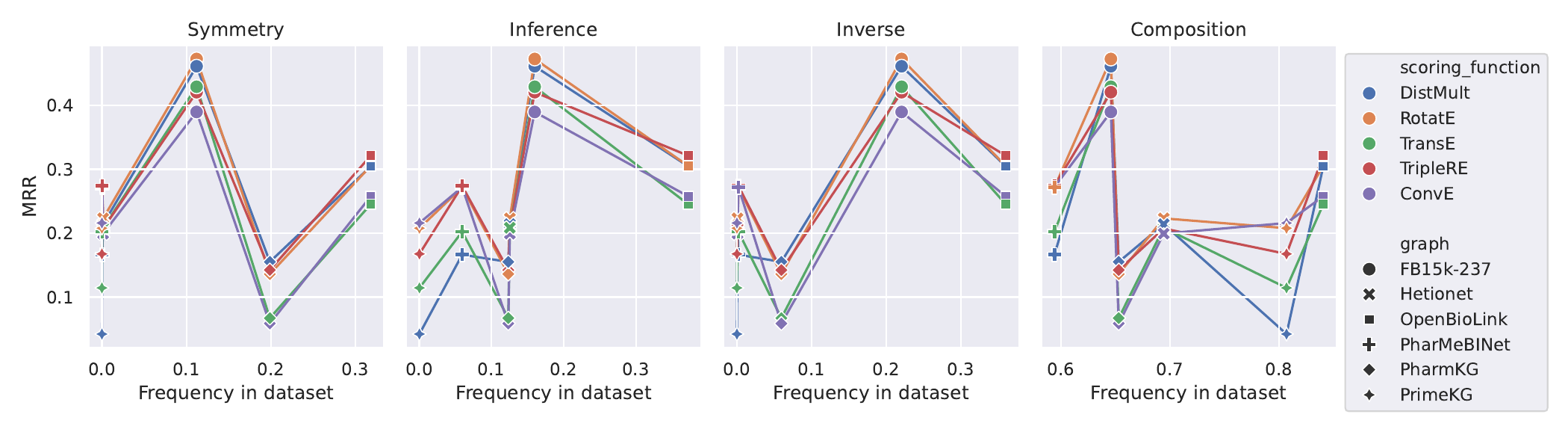}
    }
    \caption{Mean reciprocal rank (on the test set) plotted as a function of (a) the frequency of edge cardinalities (i.e., the fraction of triples in a dataset of a given edge cardinality, see \cref{fig:edge_cardinalities}) and (b) the frequency of edge topological patterns (see \cref{tab:topology_patterns}).}
\end{figure}

\begin{figure}[htb]
    \centering
    \subfloat[][\label{fig:Appendix_avg_rel_cardinalities_vs_MRR}]{
        \includegraphics[width=0.9\textwidth]{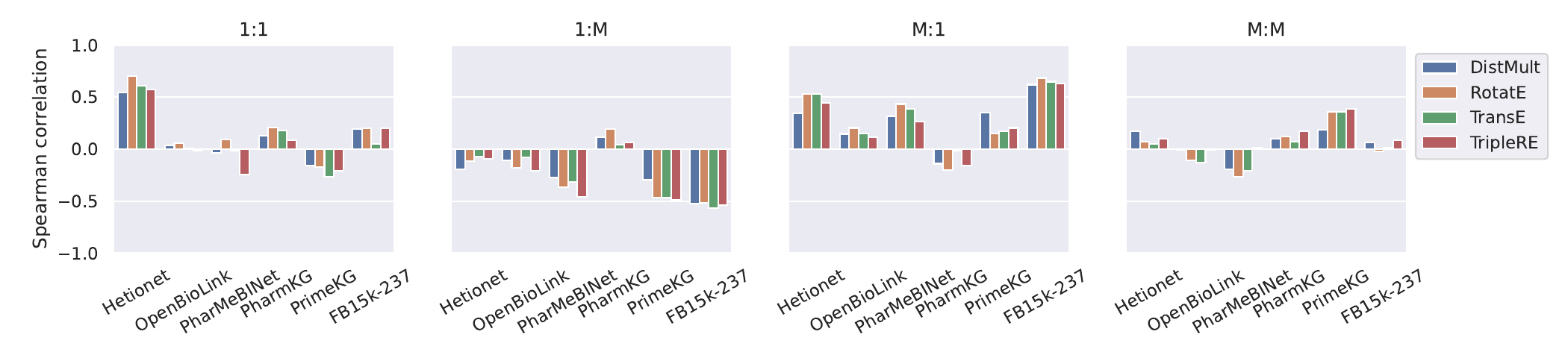}
    }\\
    \vspace{5pt}
    \subfloat[][\label{fig:Appendix_avg_rel_metrics_vs_MRR}]{
        \includegraphics[width=0.9\textwidth]{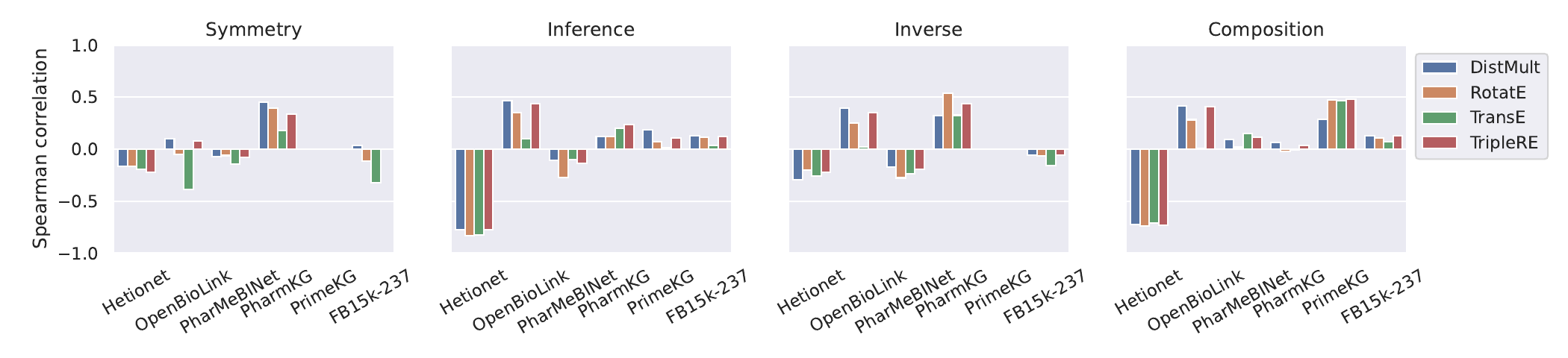}
    }
    \caption{Spearman-rank correlation between the average MRR of a relation type and the average frequency of edge cardinalities (a) and of topological patterns (b) in that relation type.}\label{fig:Appendix_avg_rel_properties_vs_MRR}
\end{figure}

\begin{figure}[htb]
    \centering
    \includegraphics[width=0.9\textwidth]{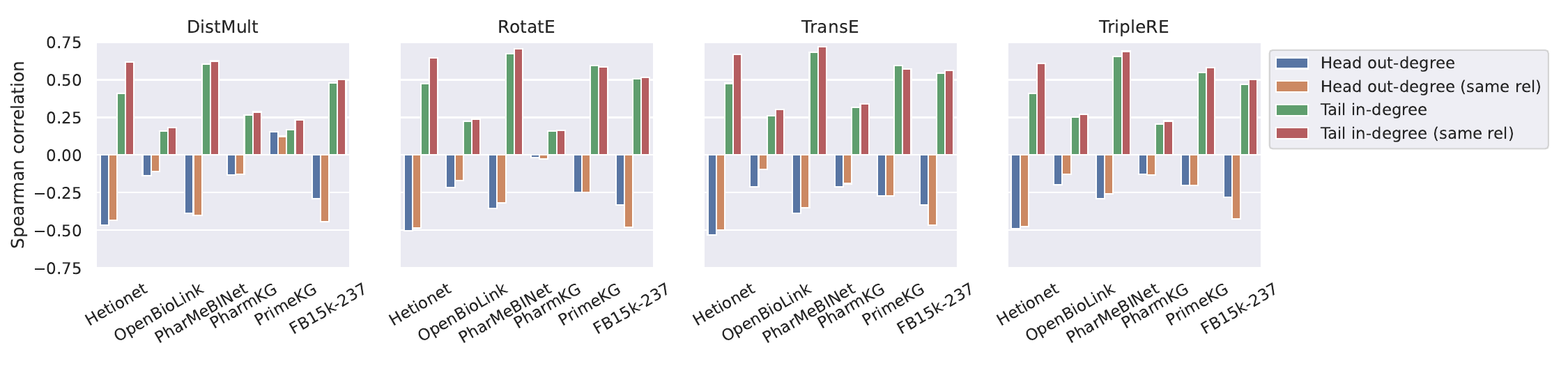}
    \caption{Spearman-rank correlation between MRR of individual triples and the out-degree of the head node as well as in-degree of the tail node.} \label{fig:Appendix_MRR_degree_correlation}
\end{figure}

\begin{figure}[htb]
    \centering
    \subfloat[][PharMeBINet]{
        \includegraphics[width=0.35\textwidth]{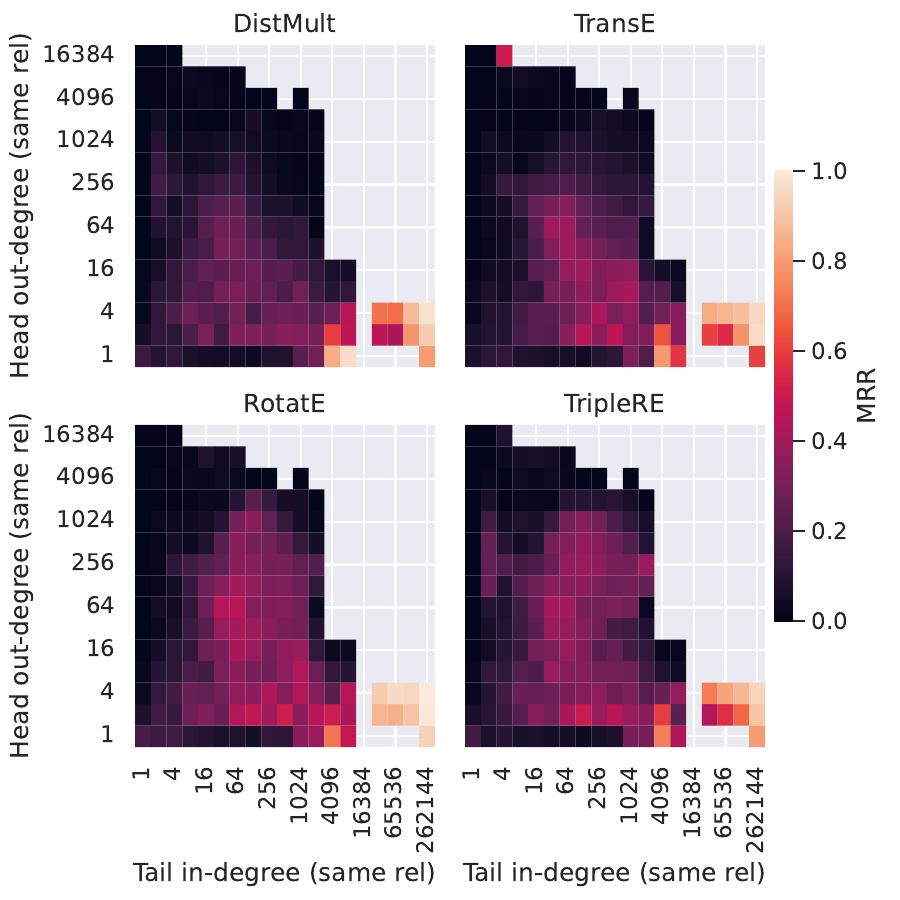}
    }
    \hspace{.025\textwidth}
    \subfloat[][PharmKG]{
        \includegraphics[width=0.35\textwidth]{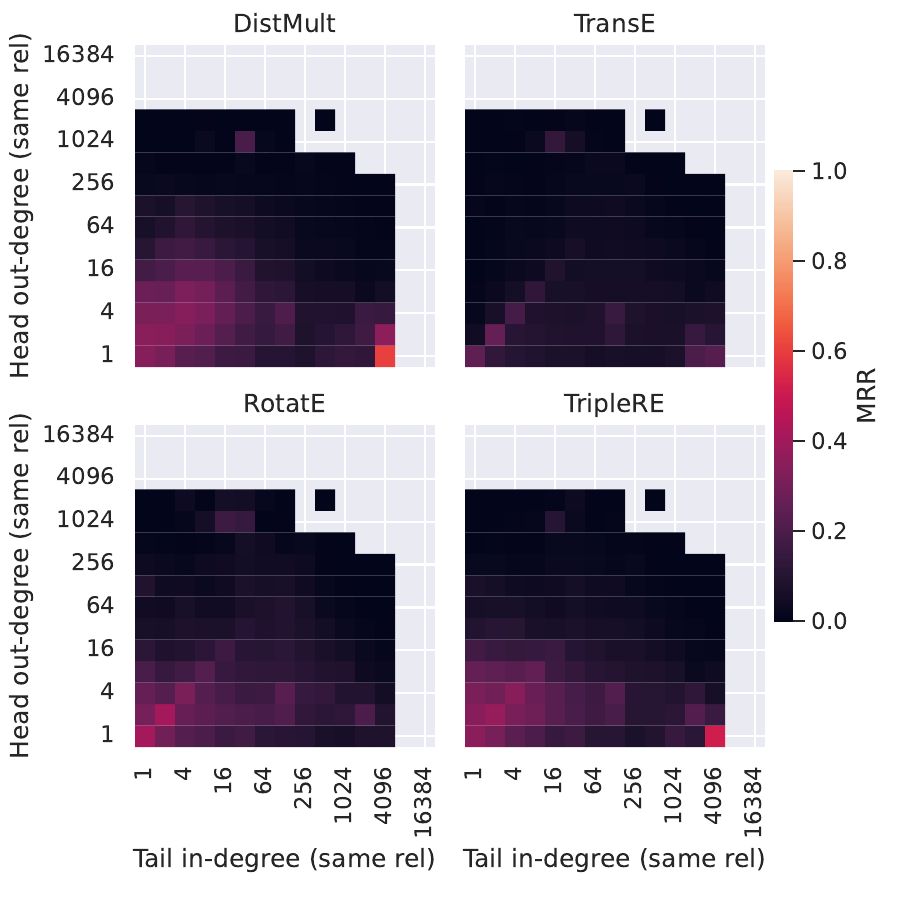}
    }

    \vspace{10pt}
    \subfloat[][PrimeKG]{
        \includegraphics[width=0.35\textwidth]{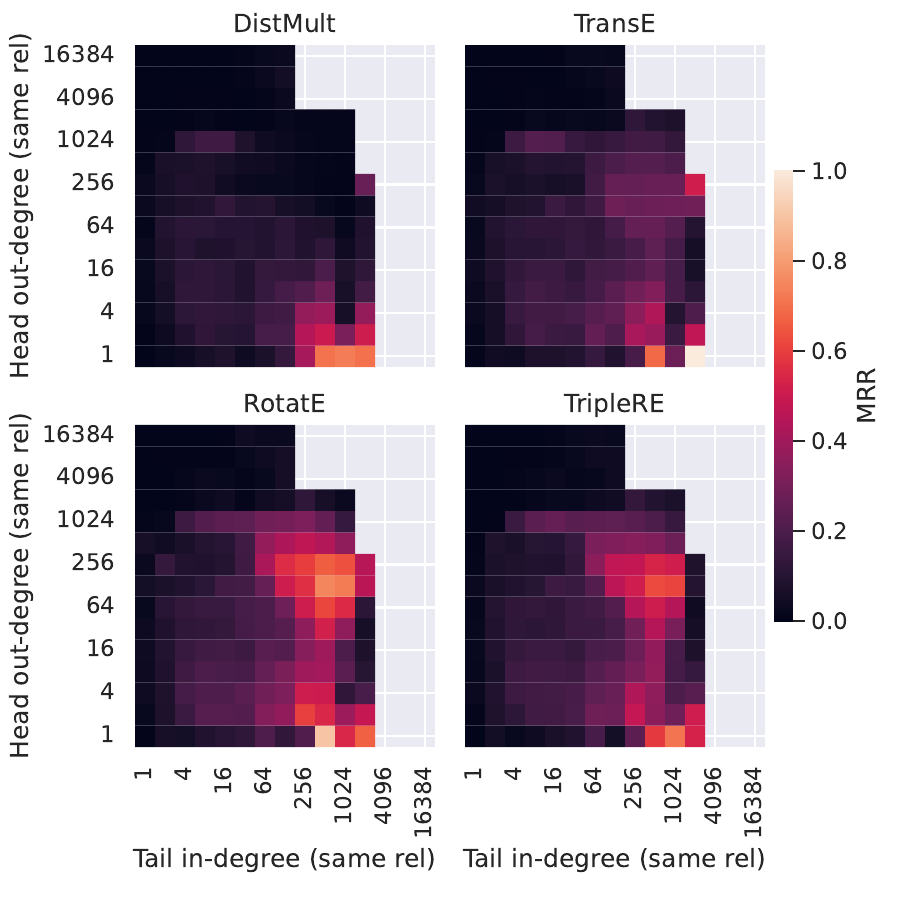}
    }
    \hspace{.025\textwidth}
    \subfloat[][FB15k-237]{
        \includegraphics[width=0.35\textwidth]{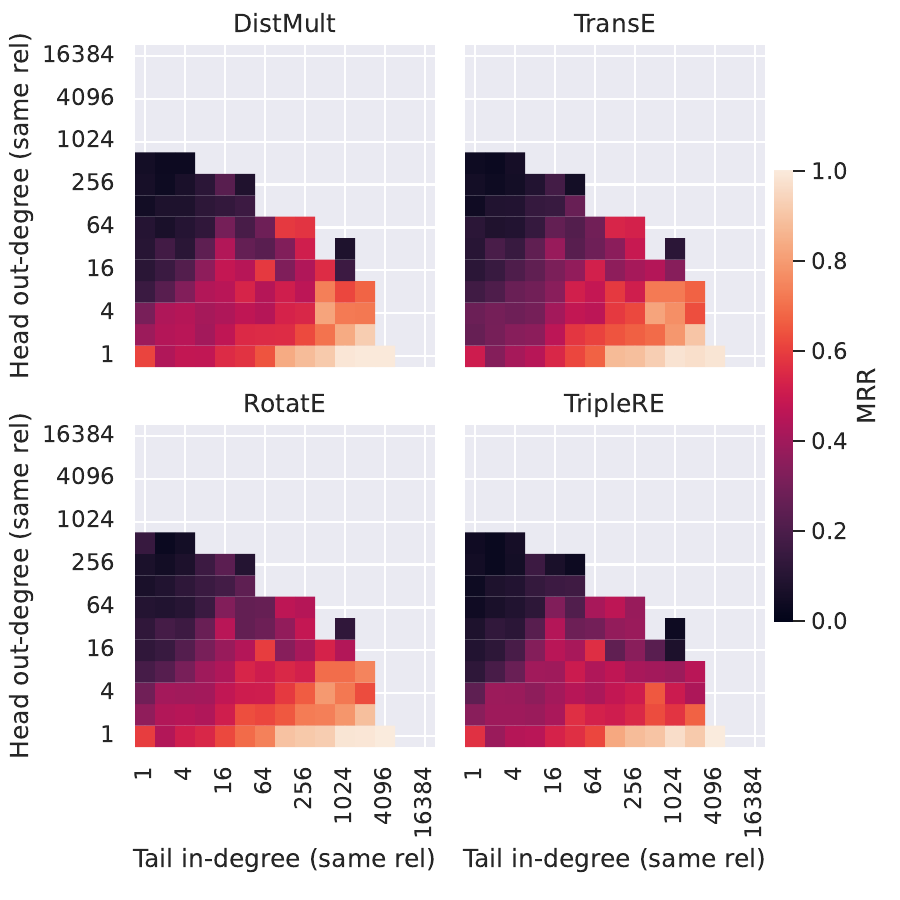}
    }
    \caption{Effect of head out-degree and tail in-degree on MRR for additional datasets.} \label{fig:Appendix_MRR_vs_degree}
\end{figure}

\begin{figure}[htb]
    \centering
    \subfloat{\includegraphics[width=0.8\textwidth]{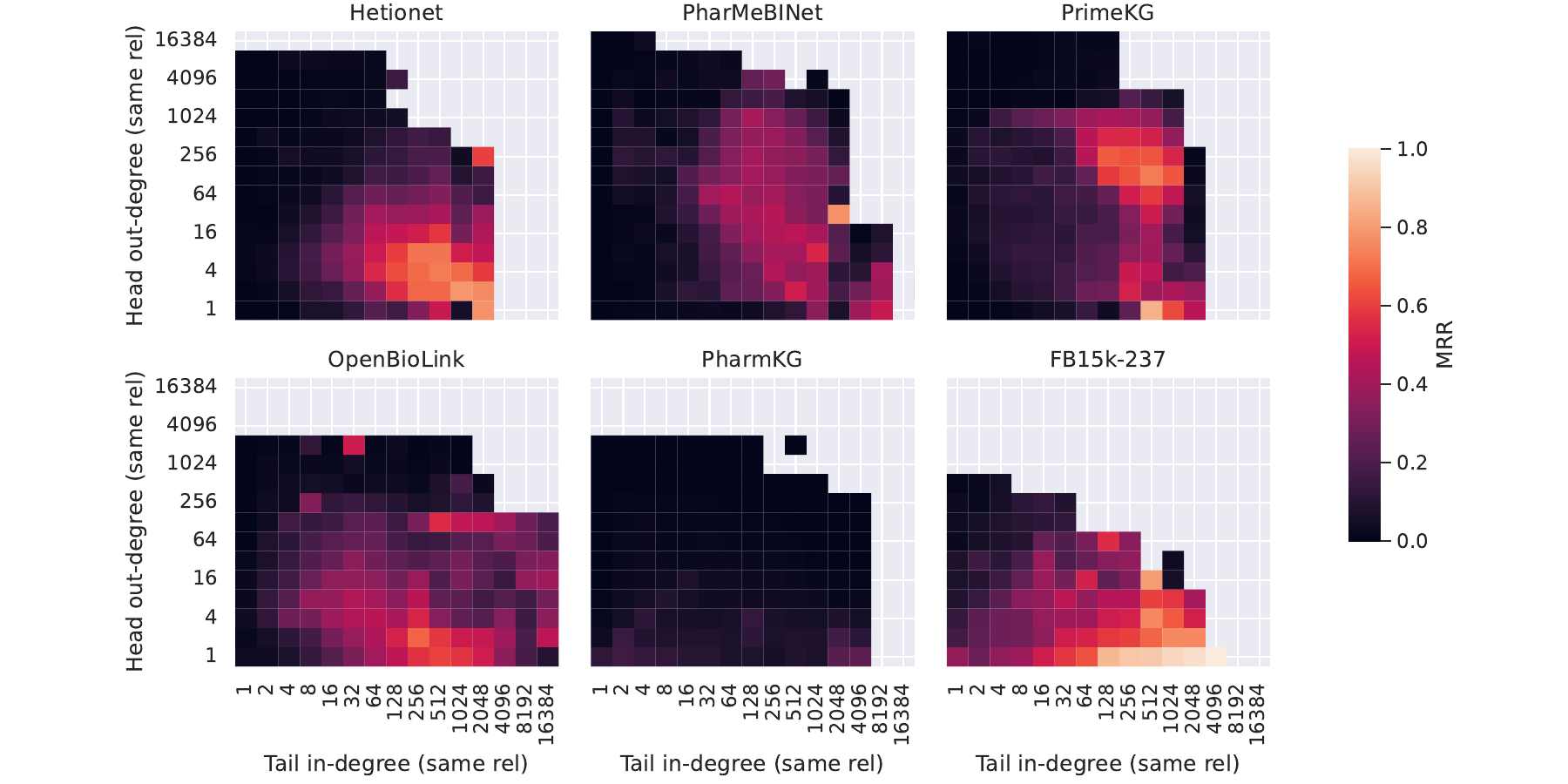}}
    \caption{Effect of head out-degree and tail in-degree on MRR, for the ConvE KGE model.}\label{fig:Appendix_conve_MRR_vs_degree}
\end{figure}

\begin{figure}[htb]
    \centering
    \subfloat{\includegraphics[width=0.35\textwidth]{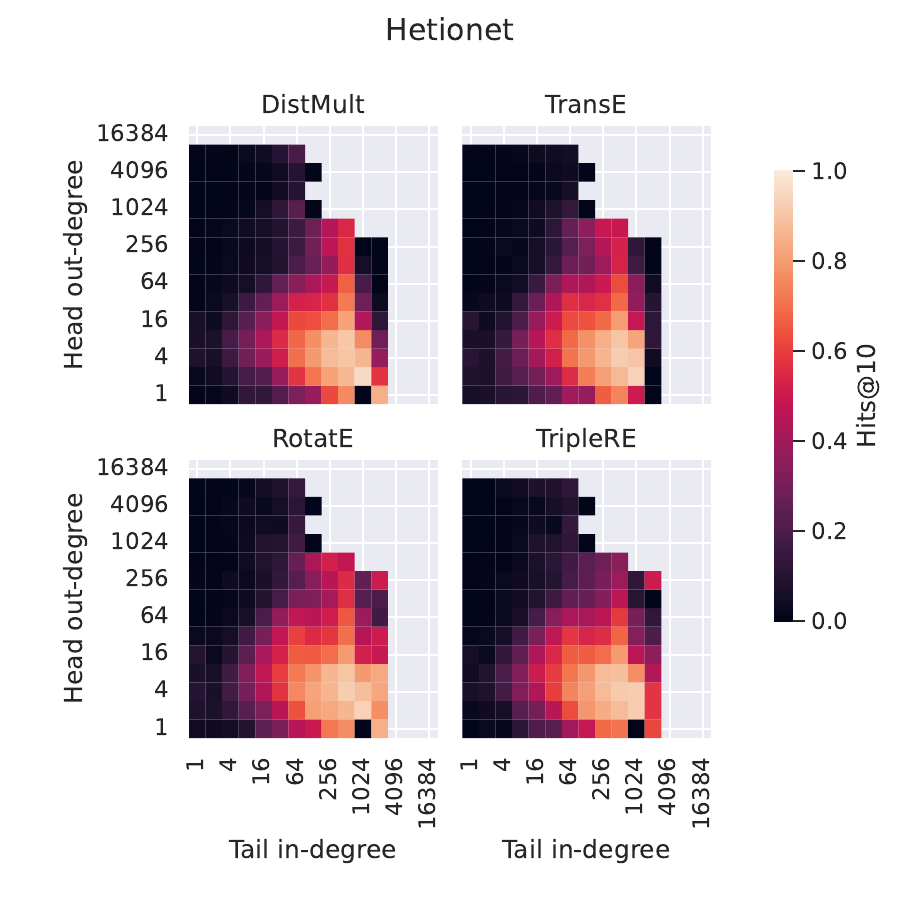}}
    \hspace{0.025\textwidth}
    \subfloat{\includegraphics[width=0.35\textwidth]{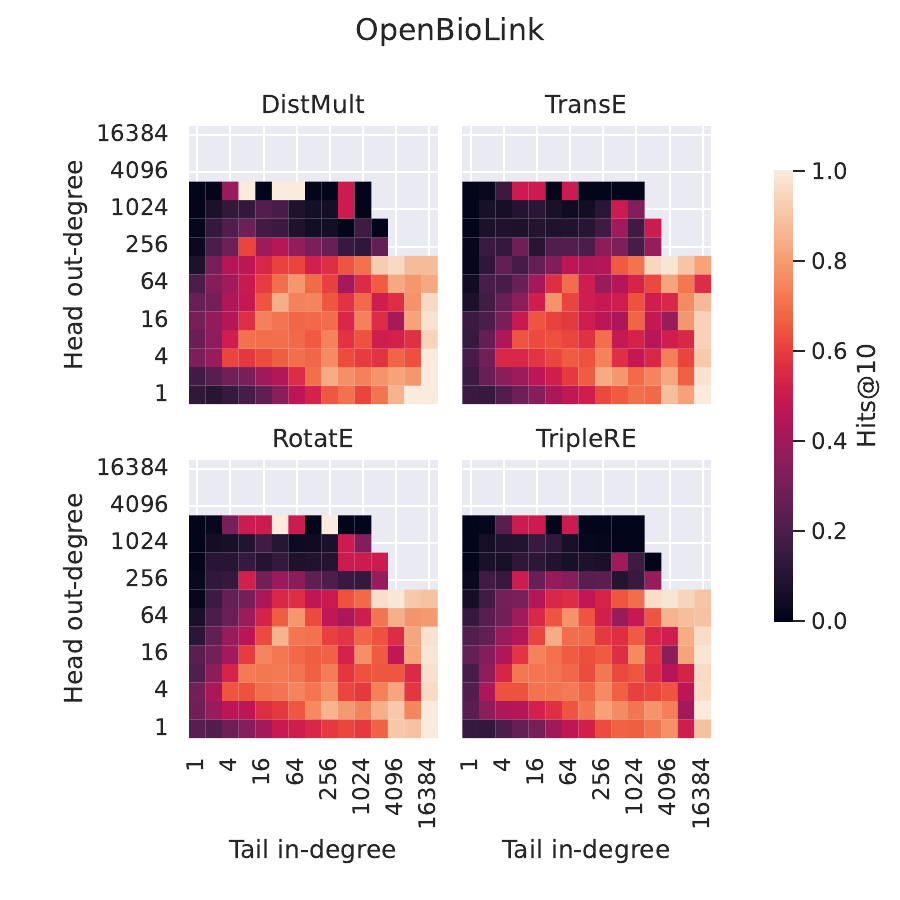}}
    \caption{Effect of the head and tail degrees of same relation type on Hits@10, for Hetionet and OpenBioLink.}\label{fig:Appendix_hits_at_10_vs_degree}
\end{figure}

\begin{figure}[htb]
    \centering
    \includegraphics[width=.55\textwidth]{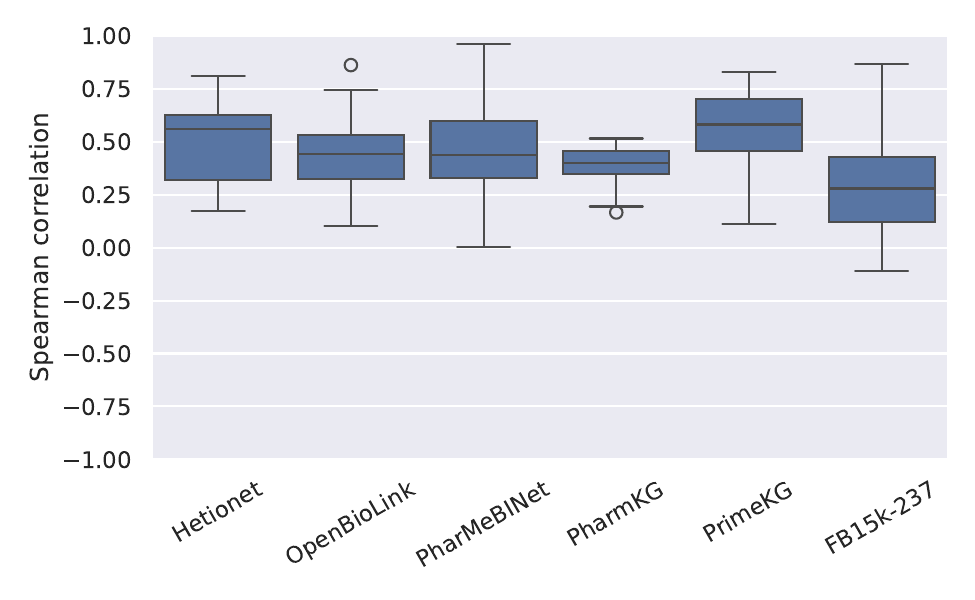}
    \caption{Distribution of Spearman-rank correlation between in-degree of same relation type and how frequently the entity is \textit{incorrectly} selected among the top-$100$ tail predictions, grouping test queries by relation type. A positive correlation means that KGE models are biased towards predicting entities with a larger number of incoming edges of the relation type considered in the query.}\label{fig:Appendix_rel_bias}
\end{figure}

\begin{figure}[htb]
    \centering
    \subfloat[][PharMeBINet]{
        \includegraphics[width=0.35\textwidth]{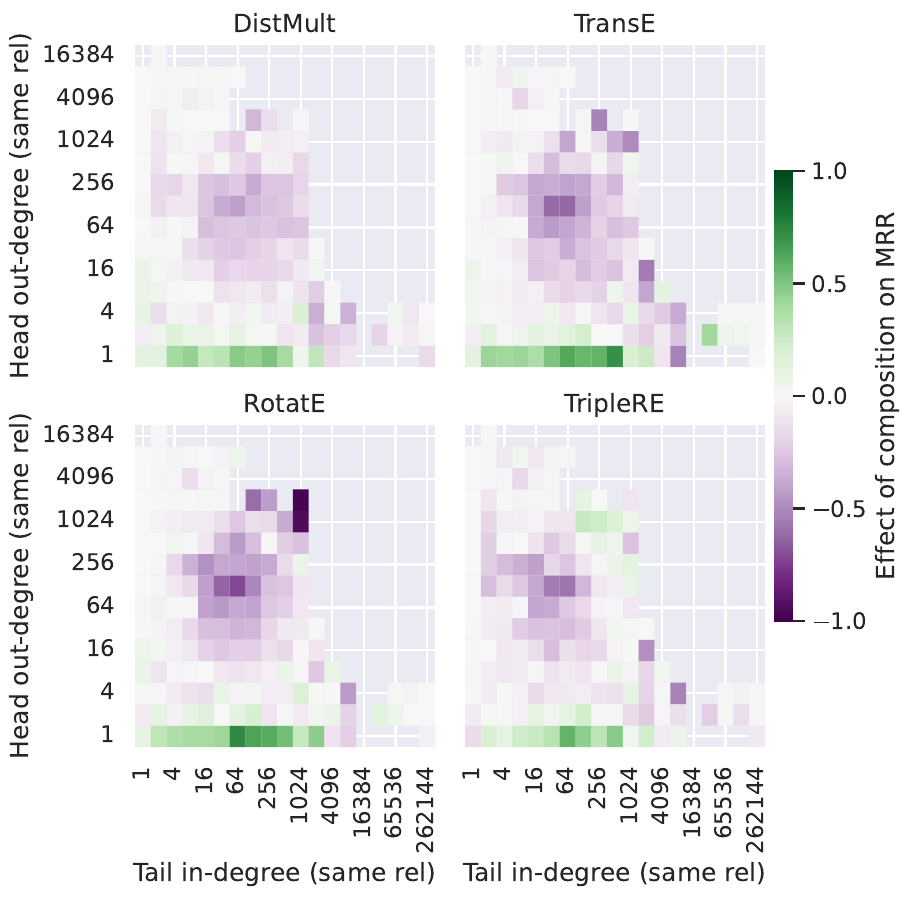}
    }
    \hspace{.025\textwidth}
    \subfloat[][PharmKG]{
        \includegraphics[width=0.35\textwidth]{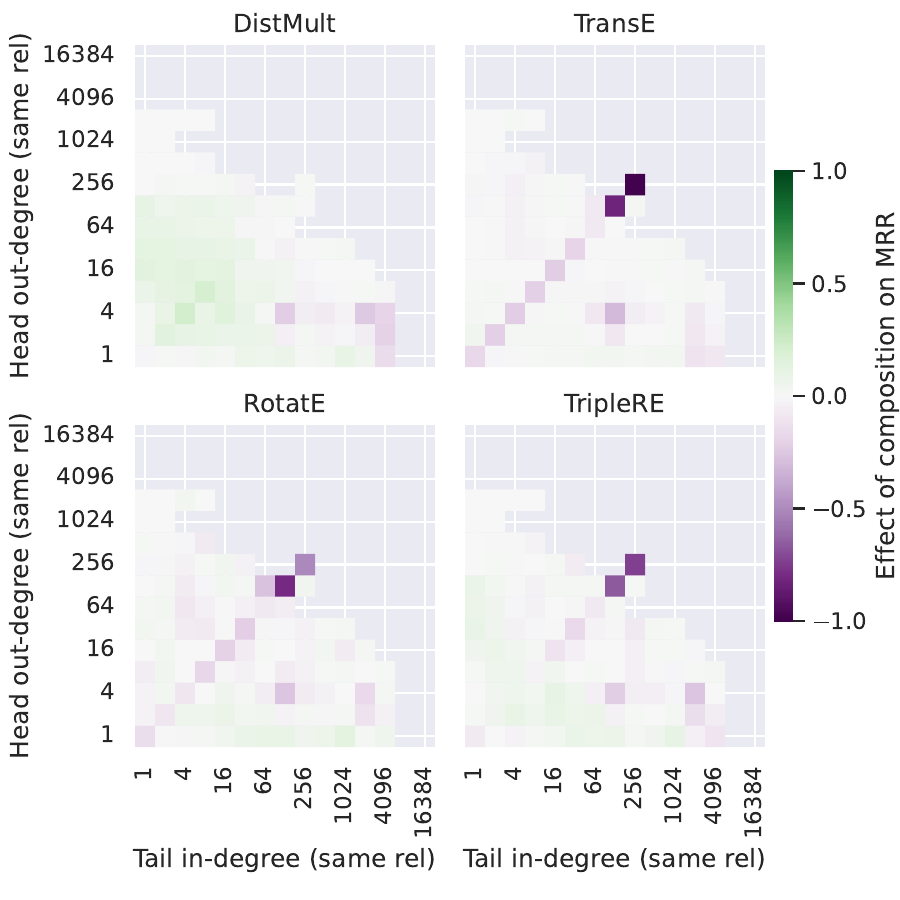}
    }

    \vspace{10pt}
    \subfloat[][PrimeKG]{
        \includegraphics[width=0.35\textwidth]{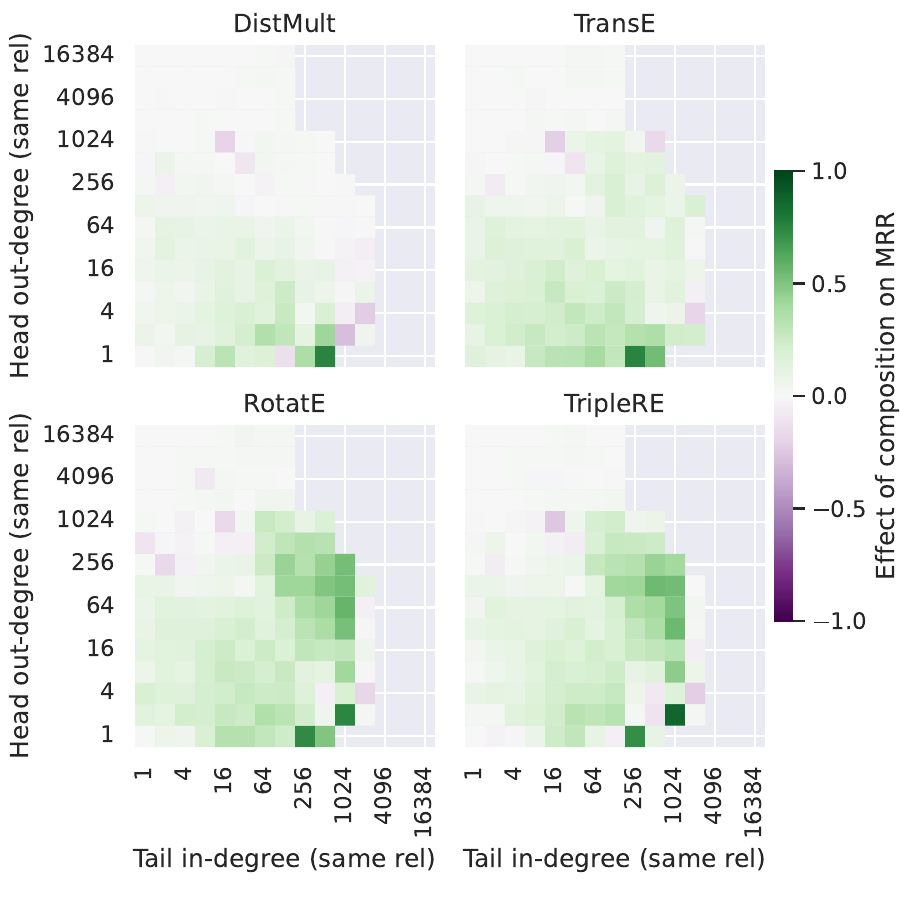}
    }
    \hspace{.025\textwidth}
    \subfloat[][FB15k-237]{
        \includegraphics[width=0.35\textwidth]{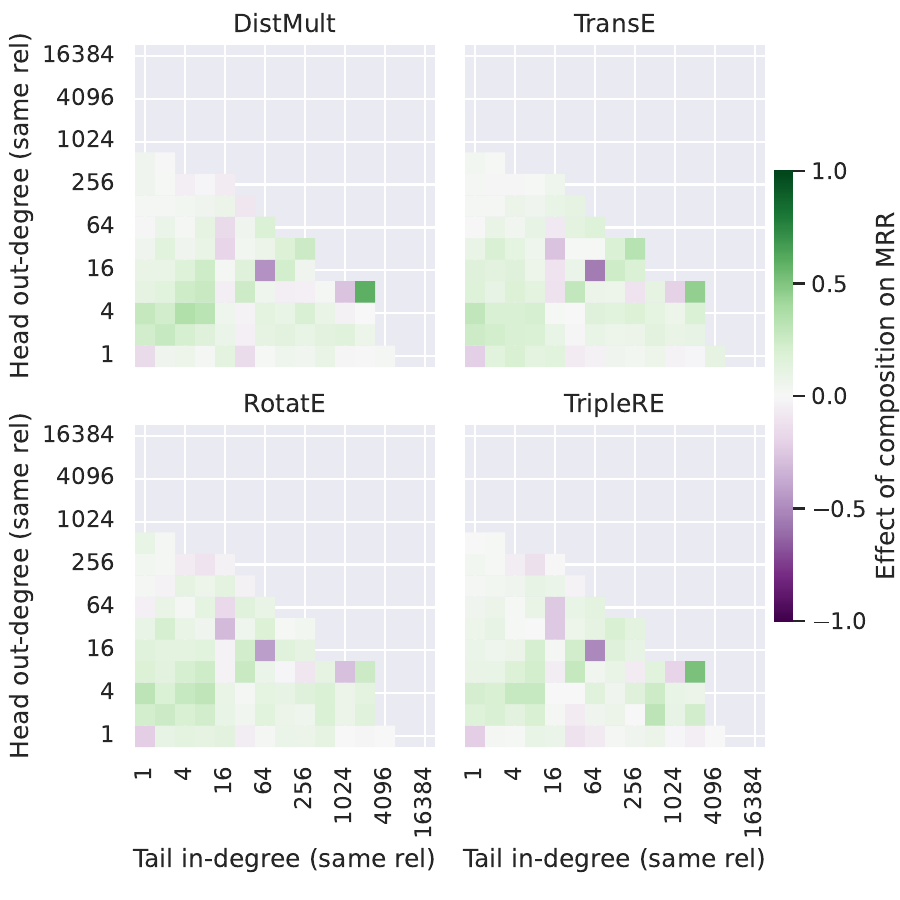}
    }
    \caption{The effect of having compositions on MRR, triples grouped by their head and tail degrees.}\label{fig:Appendix_MRR_vs_composition}
\end{figure}

\begin{figure}[htb]
    \centering
    \subfloat{\includegraphics[width=0.8\textwidth]{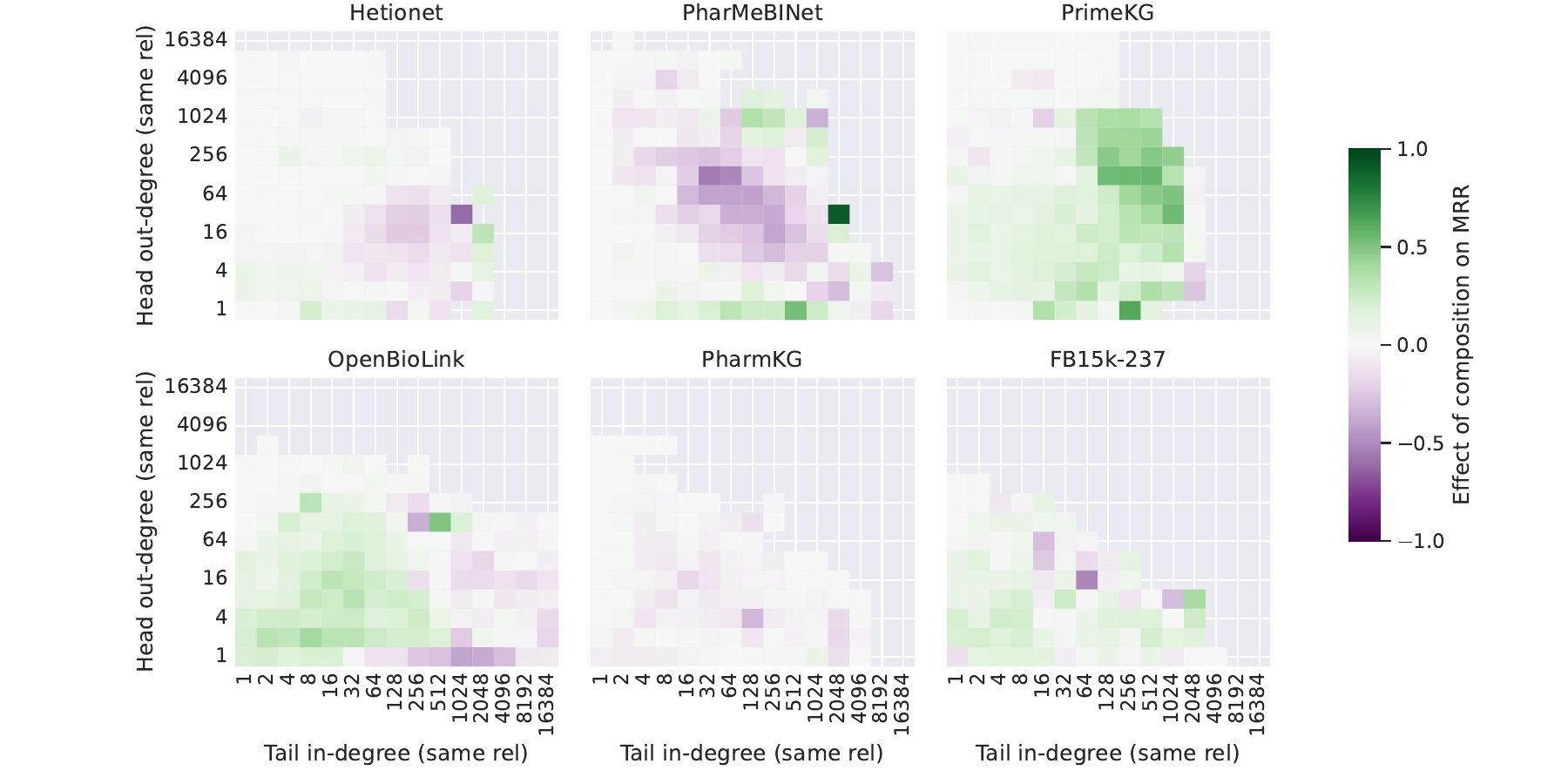}}
    \caption{The effect of having compositions on MRR, for the ConvE KGE model.}\label{fig:Appendix_conve_MRR_vs_composition}
\end{figure}

\begin{figure}[htb]
    \centering
    \subfloat{\includegraphics[width=0.35\textwidth]{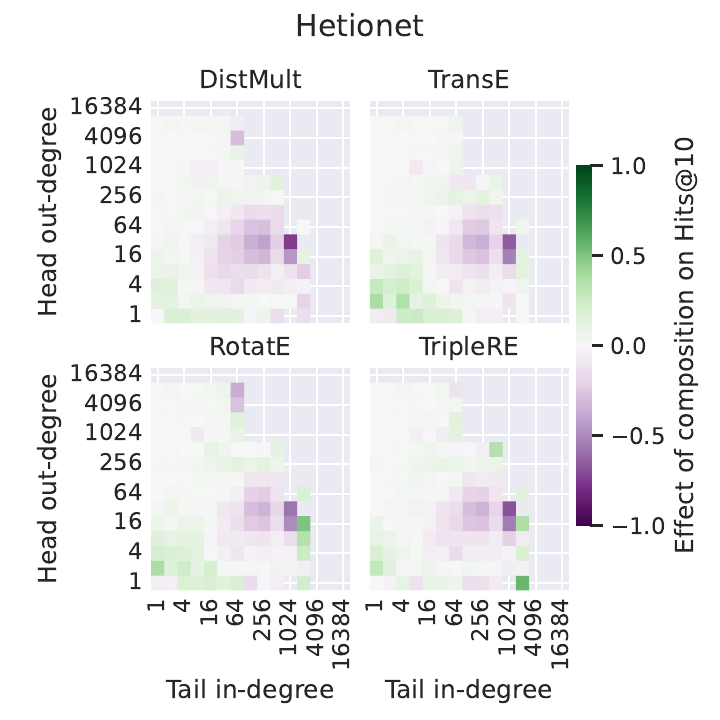}}
    \hspace{0.025\textwidth}
    \subfloat{\includegraphics[width=0.35\textwidth]{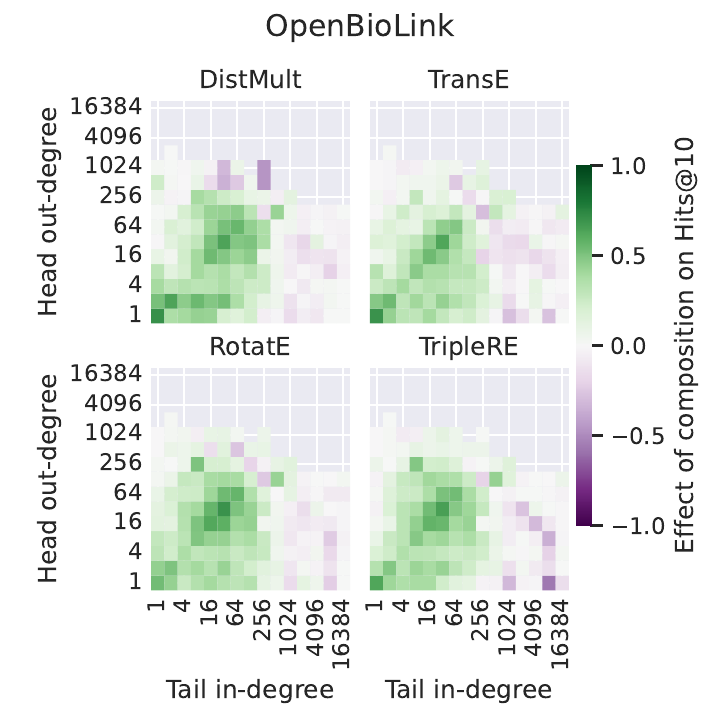}}
    \caption{The effect of having compositions on Hits@10, for Hetionet and OpenBioLink.}\label{fig:Appendix_hits_at_10_vs_composition}
\end{figure}

\begin{figure}[htb]
    \centering
    \includegraphics[height=0.9\textheight]{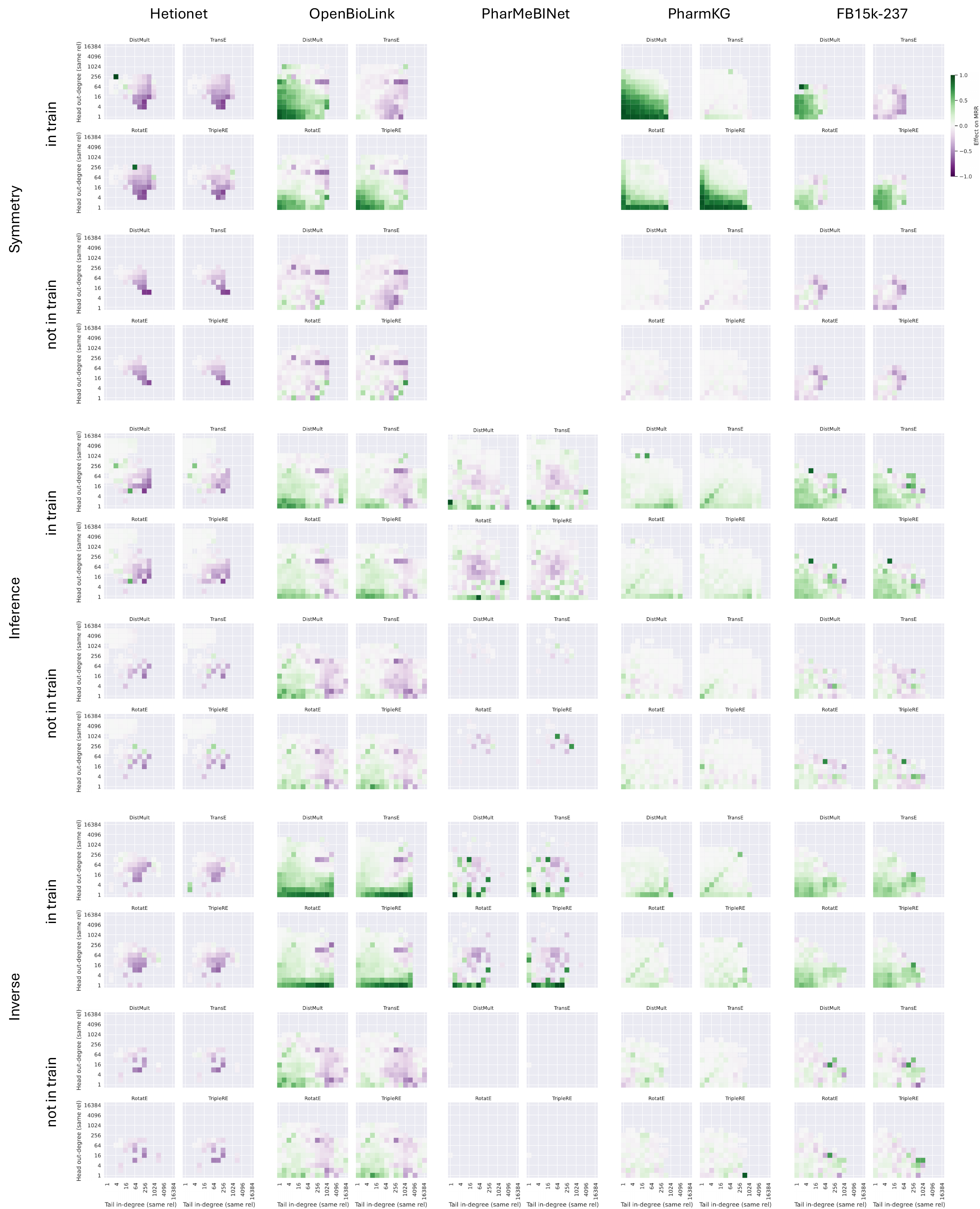}
    \caption{The effect on MRR of being symmetric and having inference/inverse, distinguishing based on whether the counterpart edge is present or absent in in the training data. Note that symmetric triples are too rare in PharMeBINet for a meaningful analysis. Triples grouped by their head and tail degrees.}\label{fig:Appendix_MRR_vs_symmetry_inference_inverse}
\end{figure}

\begin{figure}[htb]
    \centering
    \subfloat[][Median head out-degree and tail in-degree of same relation type and number of unique relation tail entities.\label{fig:interesting_relations_stats_a}]{
        \includegraphics[width=0.85\textwidth]{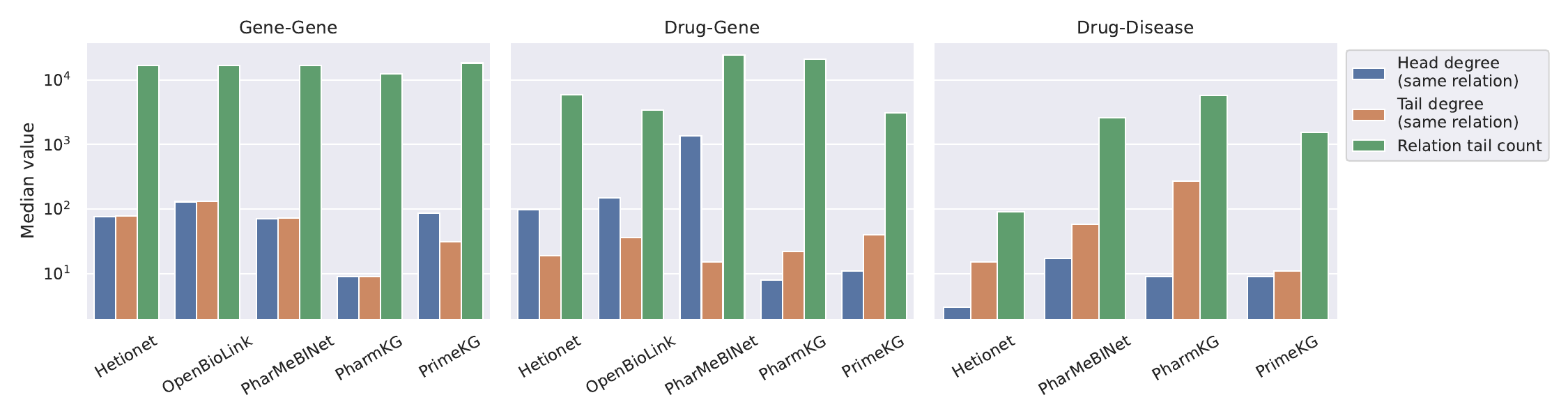}
    }

    \subfloat[][Frequency of edge patterns.\label{fig:interesting_relations_stats_b}]{
        \includegraphics[width=0.85\textwidth]{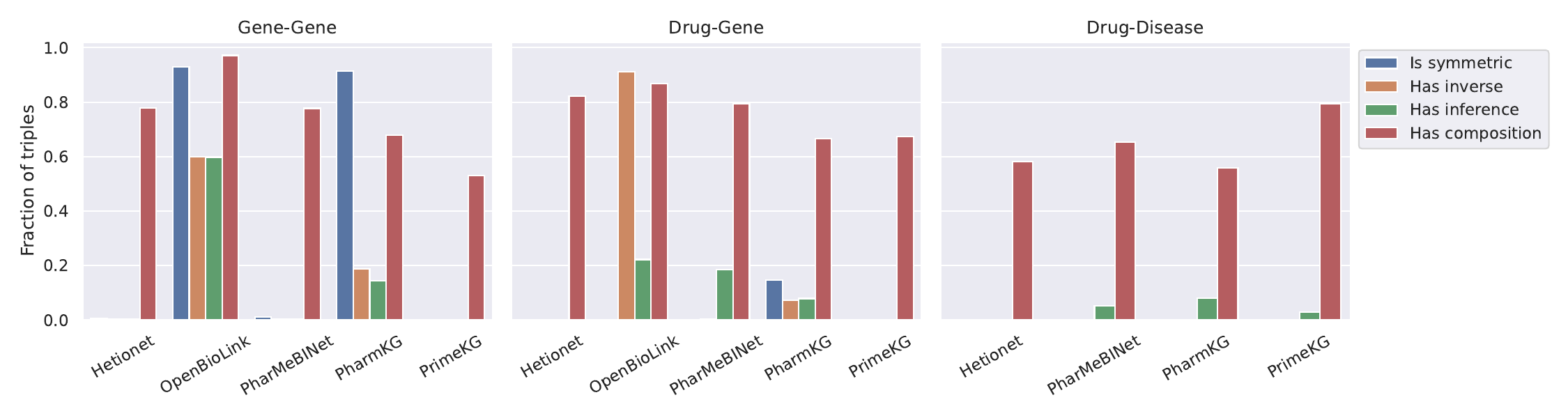}
    }
    \caption{Statistics of topological properties for different interaction types.}\label{fig:Appendix_interesting_relations_stats}
\end{figure}

\begin{figure}[htb]
    \centering
    \subfloat[][Gene-Gene\label{fig:Appendix_demixing_gg}]{
        \includegraphics[width=0.85\textwidth]{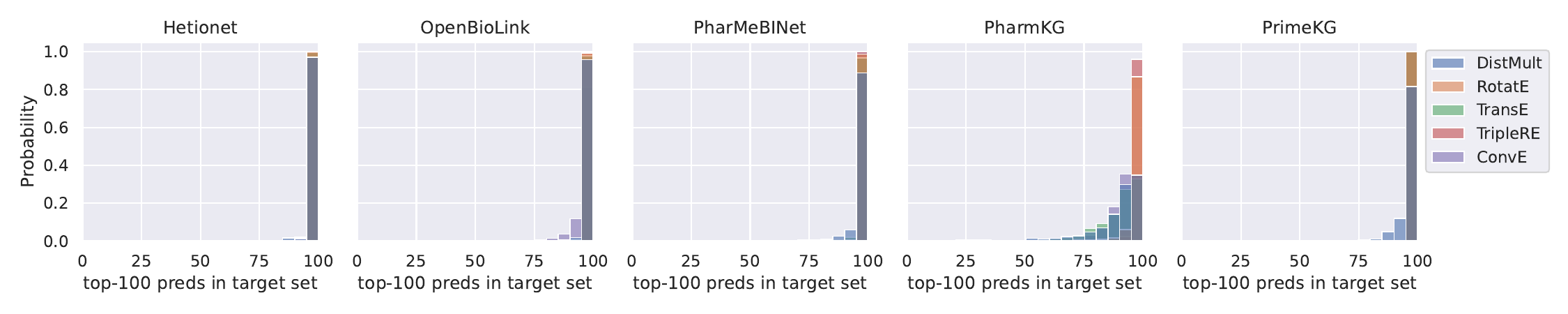}
    }

    \vspace{5pt}
    \subfloat[][Drug-Gene\label{fig:Appendix_demixing_dg}]{
        \includegraphics[width=0.85\textwidth]{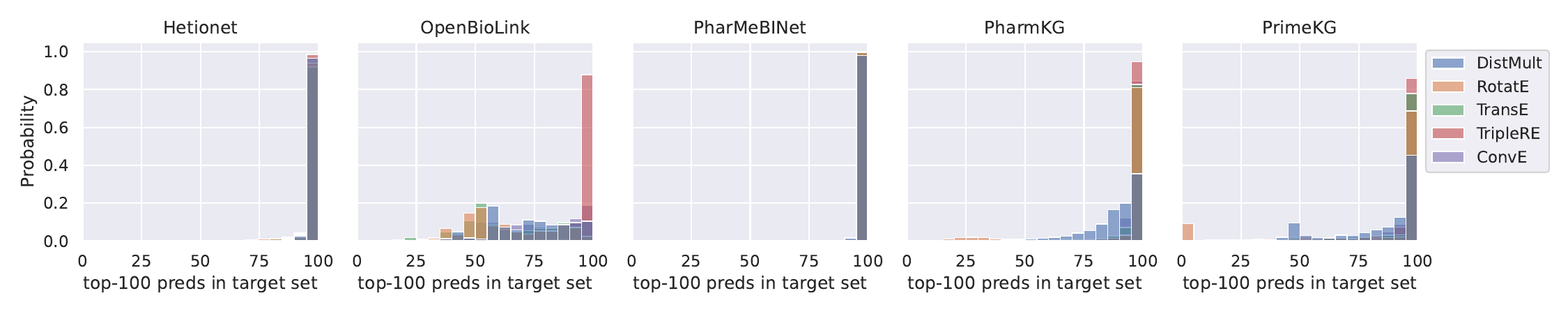}
    }

    \vspace{5pt}
    \subfloat[][Drug-Disease\label{fig:Appendix_demixing_dd}]{
        \includegraphics[width=0.7\textwidth]{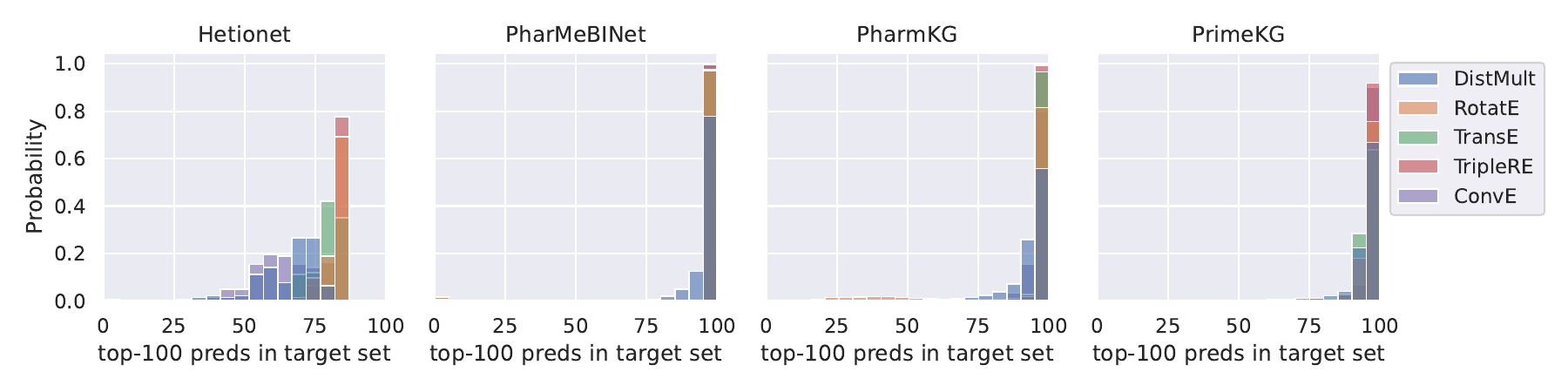}
    }
    \caption{Demixing for different interaction types. For each test query, we compute how many of the top-$100$ predictions made by the model are contained in the set of entities used as tails by triples of the considered relations.}
    \label{fig:Appendix_demixing}
\end{figure}

\begin{figure}[htb]
    \centering
    \includegraphics[width=0.9\textwidth]{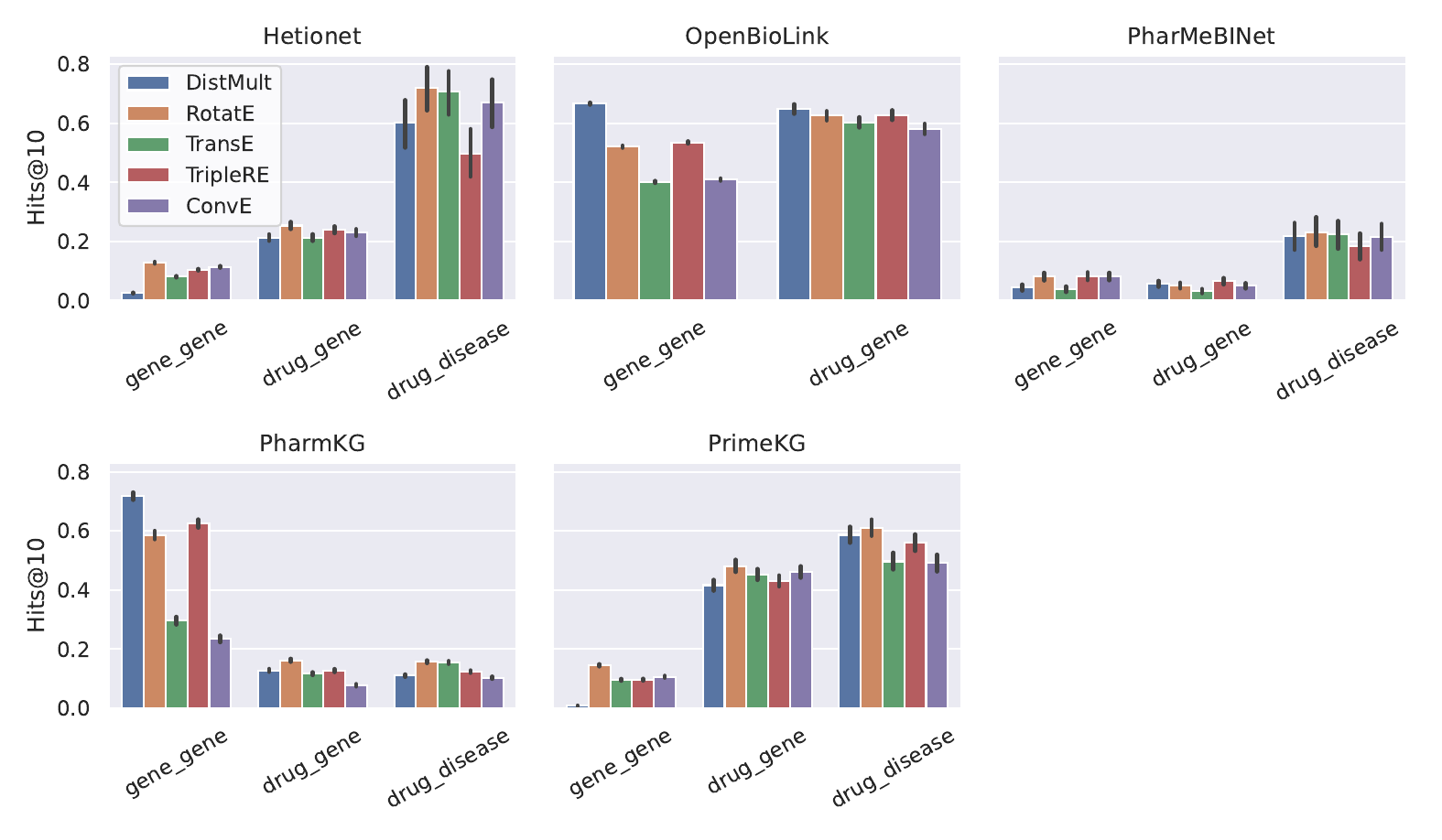}
    \caption{Hits@10 of the interaction types considered in \cref{sec:specific_relations}.}\label{fig:Appendix_hits_at_10_interesting_rels}
\end{figure}

\begin{table}[hbt]
    \scriptsize
    \renewcommand{\arraystretch}{1.3}%
    \centering
    \aboverulesep = 0pt
    \belowrulesep = 0pt
    \caption{Statistics of the relation types compared between Hetionet and PharMeBINet in \cref{subsec:hetio_phbnet_comparison}. \textit{Matching rate} denotes the fraction of triples of the relation type that are matched to triples in the other dataset. For the rows \textit{Unique heads/tails/out-relations/in-relations} and \textit{Head/Tail out-/in-degree (same relation)}, the median value is reported. For the rows \textit{Has inverse/inference/composition}, we report the fraction of total relation triples with the given property.} \label{tab:Appendix_hetio_phbnet_comparison}
    \begin{tabular}{|l|ll|ll|ll|ll|}
        \hline
        \textbf{Relation}    & \multicolumn{2}{l|}{\textbf{Disease-localizes-Anatomy}}
                             & \multicolumn{2}{l|}{\textbf{Compound-binds-Gene}}
                             & \multicolumn{2}{l|}{\textbf{Gene-covaries-Gene}}
                             & \multicolumn{2}{l|}{\textbf{Anatomy-expresses-Gene}}                         \\
        \textbf{Dataset}     & \textbf{Hetionet}                                         & \textbf{PhMBNet}
                             & \textbf{Hetionet}                                         & \textbf{PhMBNet}
                             & \textbf{Hetionet}                                         & \textbf{PhMBNet}
                             & \textbf{Hetionet}                                         & \textbf{PhMBNet} \\
        \hline
        Relation triples     & $3602$                                                    & $3602$
                             & $11571$                                                   & $11622$
                             & $61690$                                                   & $61615$
                             & $526407$                                                  & $526180$         \\
        Matching rate        & $1.0$                                                     & $1.0$
                             & $0.998$                                                   & $0.993$
                             & $0.999$                                                   & $1.0$
                             & $0.999$                                                   & $1.0$            \\
        Test triples         & $360$                                                     & $360$
                             & $1141$                                                    & $1141$
                             & $6161$                                                    & $6161$
                             & $52618$                                                   & $52618$          \\
        \hline
        Unique heads         & $133$                                                     & $133$
                             & $1389$                                                    & $1426$
                             & $9043$                                                    & $9034$
                             & $241$                                                     & $241$            \\
        Unique tails         & $398$                                                     & $398$
                             & $1689$                                                    & $1701$
                             & $9542$                                                    & $9518$
                             & $18094$                                                   & $18074$          \\
        \hline
        Head out-degree      & $212$                                                     & $227$
                             & $132$                                                     & $2085$
                             & $74$                                                      & $128$
                             & $11952$                                                   & $11945$          \\
        Head out-degree s.r. & $34$                                                      & $34$
                             & $14$                                                      & $14$
                             & $20$                                                      & $20$
                             & $7937$                                                    & $7935$           \\
        Unique out-relations & $4$                                                       & $6$
                             & $4$                                                       & $20$
                             & $4$                                                       & $8$
                             & $3$                                                       & $3$              \\
        Tail in-degree       & $11$                                                      & $11$
                             & $102$                                                     & $162$
                             & $83$                                                      & $114$
                             & $77$                                                      & $112$            \\
        Tail in-degree s.r.  & $11$                                                      & $11$
                             & $36$                                                      & $36$
                             & $17$                                                      & $17$
                             & $44$                                                      & $44$             \\
        Unique in-relations  & $1$                                                       & $1$
                             & $7$                                                       & $13$
                             & $6$                                                       & $11$
                             & $6$                                                       & $11$             \\
        \hline
        Has inverse          & $0.0$                                                     & $0.0$
                             & $0.0$                                                     & $0.03$
                             & $0.001$                                                   & $0.001$
                             & $0.0$                                                     & $0.0$            \\
        Has inference        & $0.0$                                                     & $0.0$
                             & $0.006$                                                   & $0.08$
                             & $0.002$                                                   & $0.002$
                             & $0.263$                                                   & $0.263$          \\
        Has composition      & $0.591$                                                   & $0.594$
                             & $0.571$                                                   & $0.957$
                             & $0.501$                                                   & $0.507$
                             & $0.907$                                                   & $0.907$          \\
        \hline
        \hline
        \textbf{Relation}    & \multicolumn{2}{l|}{\textbf{Compound-causes-Side Effect}}
                             & \multicolumn{2}{l|}{\textbf{Gene-regulates-Gene}}
                             & \multicolumn{2}{l|}{\textbf{Gene-interacts-Gene}}
                             & \multicolumn{2}{l|}{\textbf{Compound-downregulates-Gene}}                    \\
        \textbf{Dataset}     & \textbf{Hetionet}                                         & \textbf{PhMBNet}
                             & \textbf{Hetionet}                                         & \textbf{PhMBNet}
                             & \textbf{Hetionet}                                         & \textbf{PhMBNet}
                             & \textbf{Hetionet}                                         & \textbf{PhMBNet} \\
        \hline
        Relation triples     & $138944$                                                  & $154511$
                             & $265672$                                                  & $265667$
                             & $147164$                                                  & $147133$
                             & $21102$                                                   & $231156$         \\
        Matching rate        & $0.909$                                                   & $0.817$
                             & $0.999$                                                   & $1.0$
                             & $0.999$                                                   & $1.0$
                             & $0.997$                                                   & $0.098$          \\
        Test triples         & $12630$                                                   & $12630$
                             & $26566$                                                   & $26566$
                             & $14713$                                                   & $14713$
                             & $2105$                                                    & $2105$           \\
        \hline
        Unique heads         & $1071$                                                    & $1358$
                             & $4634$                                                    & $4634$
                             & $9526$                                                    & $9525$
                             & $734$                                                     & $2631$           \\
        Unique tails         & $5701$                                                    & $6023$
                             & $7048$                                                    & $7047$
                             & $14084$                                                   & $14073$
                             & $2880$                                                    & $21912$          \\
        \hline
        Head out-degree      & $245$                                                     & $2186$
                             & $203$                                                     & $254$
                             & $214$                                                     & $267$
                             & $515$                                                     & $4144$           \\
        Head out-degree s.r. & $201$                                                     & $182$
                             & $104$                                                     & $104$
                             & $54$                                                      & $54$
                             & $225$                                                     & $1413$           \\
        Unique out-relations & $5$                                                       & $19$
                             & $6$                                                       & $10$
                             & $5$                                                       & $10$
                             & $5$                                                       & $20$             \\
        Tail in-degree       & $164$                                                     & $254$
                             & $309$                                                     & $370$
                             & $106$                                                     & $145$
                             & $252$                                                     & $116$            \\
        Tail in-degree s.r.  & $164$                                                     & $193$
                             & $208$                                                     & $208$
                             & $27$                                                      & $27$
                             & $20$                                                      & $15$             \\
        Unique in-relations  & $1$                                                       & $3$
                             & $8$                                                       & $12$
                             & $7$                                                       & $11$
                             & $8$                                                       & $11$             \\
        \hline
        Has inverse          & $0.0$                                                     & $0.0$
                             & $0.003$                                                   & $0.003$
                             & $0.006$                                                   & $0.006$
                             & $0.0$                                                     & $0.002$          \\
        Has inference        & $0.0$                                                     & $0.142$
                             & $0.006$                                                   & $0.003$
                             & $0.006$                                                   & $0.006$
                             & $0.001$                                                   & $0.164$          \\
        Has composition      & $0.366$                                                   & $0.893$
                             & $0.881$                                                   & $0.886$
                             & $0.703$                                                   & $0.713$
                             & $0.913$                                                   & $0.800$          \\
        \hline
    \end{tabular}
\end{table}

\begin{figure}[htb]
    \centering
    \includegraphics[width=\textwidth]{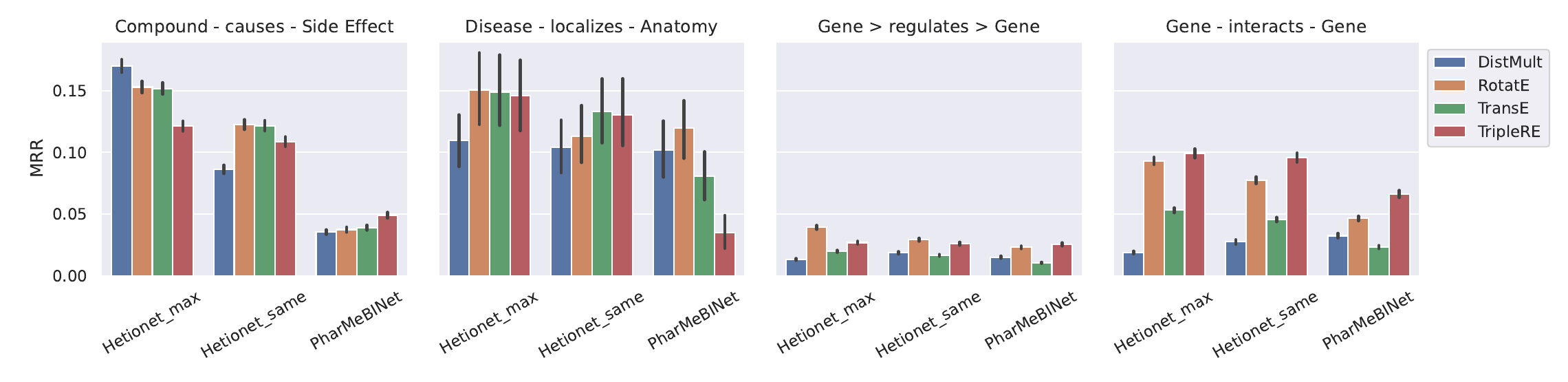}
    \caption{MRR comparison for additional relation types, when testing on a set of common edges between Hetionet and PharMeBINet.}
    \label{fig:Appendix_hetio_phbnet_other_rels}
\end{figure}

\end{document}